\pdfoutput=1

\documentclass{article} %
\usepackage{preprint,times}

\usepackage{amsmath,amsfonts,bm}

\def\eqref#1{equation~\ref{#1}}

\def\1{\bm{1}}

\DeclareMathAlphabet{\mathsfit}{\encodingdefault}{\sfdefault}{m}{sl}
\SetMathAlphabet{\mathsfit}{bold}{\encodingdefault}{\sfdefault}{bx}{n}

\usepackage{booktabs}
\usepackage{amsfonts}
\usepackage{graphicx}
\usepackage{hyperref}
\usepackage{textcomp}
\usepackage{listings}
\usepackage{adjustbox}
\usepackage{xspace}    %
\usepackage{multirow}
\usepackage{multicol}
\usepackage{xcolor} 
\usepackage{amsmath}
\usepackage{tcolorbox}
\usepackage[linesnumbered,boxed,ruled]{algorithm2e}
\usepackage{listings}
\usepackage{pifont}
\usepackage{enumitem}

\def \tool{\textsc{EvolvAttack}\xspace}

\newcommand{\fakeparagraph}[1]{\noindent\textbf{#1.}}

\usepackage{wasysym}
\usepackage{algorithmic}

\definecolor{codegreen}{rgb}{0,0.6,0}
\definecolor{codegray}{rgb}{0.5,0.5,0.5}
\definecolor{codepurple}{rgb}{0.58,0,0.82}
\definecolor{backcolour}{rgb}{0.95,0.95,0.92}

\lstdefinestyle{mystyle}{
  backgroundcolor=\color{backcolour},   commentstyle=\color{codegreen},
  keywordstyle=\color{magenta},
  numberstyle=\tiny\color{codegray},
  stringstyle=\color{codepurple},
  basicstyle=\ttfamily\footnotesize,
  breakatwhitespace=false,         
  breaklines=true,                 
  captionpos=b,                    
  keepspaces=true,                 
  numbers=left,                    
  numbersep=5pt,                  
  showspaces=false,                
  showstringspaces=false,
  showtabs=false,                  
  tabsize=2
}

\usepackage{nicematrix}
\usepackage{enumitem}

\usepackage{amsthm}
\theoremstyle{definition}

\usepackage{bbding}

\newcommand{\takeaway}[2]{%
\par\vspace{2.5mm}\noindent
\begin{tikzpicture}
\node[
  draw=black, line width=0.9pt, fill=white,
  rounded corners=1.2pt,
  inner xsep=6pt, inner ysep=6pt,
  text width=\dimexpr0.97\columnwidth-12pt\relax,
  align=left, font=\small
] (tkbody) {\rule[-0.3ex]{0pt}{3.4ex}\ignorespaces #2};
\node[
  fill=black, text=white, font=\footnotesize\bfseries,
  inner xsep=6pt, inner ysep=2.5pt,
  anchor=south west
] (tkhead) at ([xshift=4pt,yshift=0.3pt]tkbody.north west) {\strut #1};
\end{tikzpicture}\par\vspace{1.5mm}
}
\usepackage[normalem]{ulem}
\usepackage{listings}

\usepackage{hyperref}
\usepackage{url}
\usepackage{tabularx, booktabs, amssymb}
\usepackage{pgfplots}\pgfplotsset{compat=1.18}
\usepackage{float}
\usepackage[table]{xcolor}
\definecolor{headergray}{HTML}{F3F4F6}
\usepackage{booktabs, array, xcolor}
\usepackage{booktabs, pgfplots}\pgfplotsset{compat=1.18}
\usepackage{booktabs, graphicx}
\usepackage{hyperref}
\usepackage{graphicx}
\usepackage{array}
\usepackage{makecell}
\usepackage[table]{xcolor}
\usepackage{multirow}
\usepackage{pgfplots}
\pgfplotsset{compat=1.17}
\usepgfplotslibrary{groupplots}
\usepackage{subcaption}
\usepackage{wrapfig}

\usepackage{tcolorbox}
\tcbuselibrary{breakable}

\definecolor{promptblue}{RGB}{235,243,250}
\definecolor{promptframe}{RGB}{120,150,180}

\newtcolorbox{promptbox}[1]{
    colback=promptblue,
    colframe=promptframe,
    boxrule=0.45pt,
    arc=1.5pt,
    left=5pt,
    right=5pt,
    top=4pt,
    bottom=4pt,
    title=\textbf{#1},
    coltitle=black,
    fonttitle=\small,
    colbacktitle=promptblue,
    breakable,
    before skip=5pt,
    after skip=7pt
}

\title{\tool: Backdooring self Evolving VLM Agents}

\preprintfinalcopy 

\author{Chuan Zou \\
University of Pennsylvania\\
\And
Yang Xiang\\
Brown University\\
\And
Zexin Li\\
Nanyang Technological University\\
\And
Xiaonan Luo\\
University of Notre Dame\\
\And
Simin Chen\thanks{Corresponding author.}\\
George Mason University\\
}

\begin{document}

\maketitle
\vspace{-16pt}
\lhead{Preprint}  

\begin{abstract}
Self-evolving mechanisms let vision-language model (VLM) agents continually adapt through evolving memory, but they also allow malicious behavior to emerge only after deployment.
We introduce a new attack surface, the \emph{self-evolving backdoor}, whose payload stays dormant in the initial agent and is progressively activated as the agent evolves.
Unlike conventional backdoors, which can be exposed by testing triggered inputs on a static model, our attack requires both a visual trigger and a sufficiently evolved memory state; static scanning with clean memory thus yields near-zero attack success and may deem the agent benign.
We realize this through memory-aware backdoor training, which couples trigger activation with the agent's evolving memory while preserving benign behavior unless the trigger co-occurs with payload-related memory.
Across three VLM backbones, four agent pipelines, and recency-, semantic-, and reward-based memory evolution, the attack preserves normal functionality and stays nearly inactive before evolution (0.5\% attack success rate), yet exceeds 95\% attack success within 100 interactions, revealing a stealthy threat unique to self-evolving agents.
\end{abstract}

\section{Introduction}
\label{sec:introducion}

VLM agents integrate visual perception and language reasoning to act in interactive environments~\cite{zheng2024seeact,hong2024cogagentvisuallanguagemodel,he2024webvoyager,driess2023palmeembodiedmultimodallanguage}.
Self-evolving mechanisms further let them continually update their memory from prior interactions and adapt future decisions accordingly~\cite{jin2026sega,sun2026magnet,park2023generativeagents,wang2023voyager,shinn2023reflexion,xu2025amem}.

However, as agents evolve through long-term interaction, their security can no longer be judged solely at initial deployment~\cite{shao2025misevolve,zhao2026safetyrisk}.
This introduces a new attack surface, the \emph{self-evolving backdoor}: unlike conventional backdoors activated by a triggered input alone, it requires both a triggered input and a sufficiently evolved memory state.
It therefore stays dormant during pre-deployment evaluation, even on triggered inputs, and can evade conventional security checks.
This stealth does not rely on hiding the deployment memory: the attacker only poisons a small fraction of the fine-tuning data, and the activating memory is written back by the agent itself during deployment. Thus, even an evaluator using a mock or distribution-matched memory observes nearly benign behavior; exposing the backdoor requires replaying a long, trigger-bearing interaction trajectory with memory write-back.
For instance, an embodied assistant may pass initial security checks and behave normally in early use, yet execute attacker-specified actions once sufficient interaction history accumulates.
Benign behavior under static evaluation therefore no longer guarantees agent security, raising a fundamental research question: 

\takeaway{Fundamental research question}{Can a backdoored model remain benign even on triggered inputs during initial evaluation, yet reliably activate through a VLM agent's memory evolution while preserving normal functionality?}

Existing backdoor attacks target \textit{static} models~\cite{liang2024badclip,lyu2024trojvlm,yuan2025badtoken, lyu2025vlood,liu2024contextual,cheng2025ghosthand}: they bind the trigger directly to the payload, so malicious behavior activates immediately regardless of memory and can be exposed by pre-deployment trigger scanning.
Exploiting self-evolution instead requires \emph{delayed trigger activation}: the agent must behave benignly on triggered inputs during initial evaluation, yet reliably turn malicious as its memory evolves.

This poses three challenges. (1) \emph{Joint activation}: malicious behavior should emerge only when both the trigger and a sufficiently evolved memory state are present. (2) \emph{Dynamic memory}: the attack cannot rely on a fixed memory snapshot or a predetermined interaction sequence, since memory evolves across interactions. (3) \emph{Functionality preservation}: the agent must behave normally whenever the trigger and payload-related memory are not jointly present.

Our key insight is that an agent's evolving memory can serve as a latent activation condition for a visual backdoor.
Rather than binding the trigger directly to the payload, we train the model under simulated memory states at different evolution stages, so that the trigger elicits the payload only when the retrieved memory contains payload-related experiences. Memory evolution thus becomes part of the activation mechanism: ordinary interactions progressively drive the agent toward the malicious state, while the payload stays nearly inactive before evolution.

Based on this insight, we introduce \tool, a memory-aware backdoor attack against self-evolving VLM agents. The backdoored agent behaves normally when its memory is clean or insufficiently evolved, even under the trigger; as memory evolves, the trigger increasingly elicits the attacker-specified behavior. \tool thus exploits the agent's intended adaptation process while remaining benign unless the trigger co-occurs with payload-related memory.

We evaluate \tool across three VLM backbones, four agent pipelines, and recency-, semantic-, and reward-based memory evolution. Before evolution, the attack stays nearly dormant (\(0.5\%\) attack success rate) while preserving normal functionality; within 100 interactions, its attack success rate exceeds \(95\%\). Thus, an agent can appear benign even when tested with the correct trigger, yet become reliably backdoored as its memory evolves, showing that evaluating self-evolving agents only at initialization is insufficient and motivating testing across full interaction trajectories.

Our contributions are summarized as follows: %

\begin{itemize}[leftmargin=10pt]
    \item \textbf{New Attack Surface.}
    To our knowledge, we are the first to identify and formulate the \emph{self-evolving backdoor}, a new attack surface of self-evolving VLM agents. Its activation requires both a visual trigger and a sufficiently evolved memory state, so it can evade static pre-deployment evaluation even on triggered inputs.

    \item \textbf{Memory-Aware Attack.}
     We introduce \tool, a memory-aware backdoor training algorithm that couples trigger activation with the agent's evolving memory while remaining benign unless the trigger co-occurs with payload-related memory. It trains the model under simulated memory states with four complementary losses and bootstraps the first payload write-back, so that the agent's standard memory update amplifies the attack without any post-deployment parameter update.

    \item \textbf{Comprehensive Evaluation.}
    We evaluate \tool on three VLM backbones (Qwen2.5-VL-3B, InternVL3-2B, and DeepSeek-VL-1.3B) and four agentic pipelines (AppAgent, SeeAct, Mobile-Agent, and Agent-S), showing that the attack generalizes across model architectures and memory-evolution mechanisms. It stays nearly dormant before evolution (\(0.5\%\) ASR), exceeds \(95\%\) ASR within 100 interactions while keeping clean accuracy within 3\% of the clean baseline, and survives one-shot memory sanitization and dilution.
\end{itemize}

\section{Background \& Related Work}
\label{sec:background}
\fakeparagraph{VLM Agents}
Vision-Language Models (VLMs) have demonstrated strong capabilities in cross-modal reasoning and instruction following, enabling a wide range of agent applications~\cite{zheng2024seeact,gupta2023,suris2023vipergpt,%
liu2024llavaplus,driess2023palmeembodiedmultimodallanguage,zitkovich2023rt2}.
When deployed as interactive agents, VLMs perceive visual observations and
issue executable actions to accomplish goals.
Formally, at step~$t$ of agent execution, a VLM $F_\theta$ receives an
observation $I_t$, an instruction $q_t$, and a context $C_t = h(y_1,\ldots,y_{t-1})$ built from its previous outputs by a context-construction function $h(\cdot)$, and produces an output
$y_t = F_\theta(I_t, q_t, C_t)$. 
This paradigm has been validated across diverse settings, including web
navigation~\cite{zheng2024seeact}, embodied
manipulation~\cite{driess2023palmeembodiedmultimodallanguage,zitkovich2023rt2}, and visual
reasoning~\cite{gupta2023,suris2023vipergpt}.

\fakeparagraph{Self-Evolving Agents with Memory Updates}
In this work, we consider a VLM $F_\theta(\cdot)$ deployed within an
evolving agent pipeline that maintains a memory state $M_t$ across
interactions. At step $t$, the agent retrieves task-relevant context
$C_t = f_{\mathrm{ret}}(M_t, q_t)$, where $f_{\mathrm{ret}}(\cdot)$
denotes the memory retrieval module, and generates
$y_t = F_\theta(I_t, q_t, C_t)$. Since the memory is updated after every step, $M_t = f_{\mathrm{upd}}(M_{t-1}, y_{t-1})$ already incorporates the previous output; hence $y_t$ is conditioned on $y_{t-1}$, and on earlier outputs, through the retrieved context $C_t$.
The resulting trajectory is then incorporated into memory through
$M_{t+1} = f_{\mathrm{upd}}(M_t, y_t)$, where
$f_{\mathrm{upd}}(\cdot)$ denotes the memory update module.
Through this retrieve-interact-update loop, the agent continuously evolves
its memory over time. 
When deployed within a memory-augmented pipeline, a VLM maintains a
persistent memory pool $M_t$ across interactions~\cite{park2023generativeagents,
wang2023voyager,zhao2024expel,zhong2024memorybank,xu2025amem}.
At each step~$t$, a retrieval operator $f_{\mathrm{ret}}$ selects a context
$C_t \subseteq M_t$ according to one of three common
paradigms: \emph{recency-based} retrieval returns the last $k$ entries,
as in MemGPT~\cite{packer2023memgpt} and Generative
Agents~\cite{park2023generativeagents}; \emph{embedding-based} retrieval
returns the top-$k$ entries by similarity to the current query, as in
A-MEM~\cite{xu2025amem} and Reflexion~\cite{shinn2023reflexion}; and
\emph{reward-weighted} retrieval ranks entries by both similarity and
reward, as in Voyager~\cite{wang2023voyager} and ExpeL~\cite{zhao2024expel}. The VLM then generates
$y_t = F_\theta(I_t,q_t,C_t)$, and the resulting trajectory is written
back to memory through an update operator $f_{\mathrm{upd}}$, typically
implemented as a gated-append rule~\cite{zhang2023appagent,
shinn2023reflexion}.

\fakeparagraph{Backdoor Attacks and Defenses for VLM Agents}
Backdoor attacks implant hidden malicious behaviors during training, causing
a model to behave normally on clean inputs but produce attacker-specified
outputs when a trigger is present. In VLMs, prior work has studied
parameter-level backdoors induced by visual triggers on both contrastive and
generative models~\cite{liang2024badclip,lyu2024trojvlm,lyu2025vlood,
liu2024contextual,yuan2025badtoken}.
Dual-Key Backdoors and Composite Backdoor Attacks further study backdoors
that activate only under multiple trigger conditions~\cite{walmer2022dualkeymultimodalbackdoorsvisual,
huang2024compositebackdoorattackslarge}, while Sleeper Agents show that malicious behavior can remain
dormant until a specific deployment context is encountered~\cite{hubinger2024sleeperagentstrainingdeceptive}.
 while BackdoorAgent~\cite{feng2026backdooragent}
extends backdoor analysis to agent components such as planning, memory, and
tool use. 
Recent work further targets memory-augmented agents.
AgentPoison~\cite{chen2024agentpoison} and MINJA~\cite{dong2025memory}
manipulate retrieved memory through direct poisoning or query-only interaction,
while MemVenom~\cite{zhang2026memvenom} and LUCID~\cite{bouzidi2026lucid}
exploit visual inputs to contaminate or attack long-term memory.
Other studies investigate persistent or delayed threats in evolving
agents~\cite{yang2026zombie,li2026sleeper,pulipaka2026sleepermem,
wang2026oep}, while a recent survey systematizes security threats across
the memory lifecycle~\cite{lin2026memsecsurvey}.

\begin{figure}[t]
\centering
\begin{subfigure}[t]{0.48\textwidth}
  \centering
  \includegraphics[width=\linewidth]{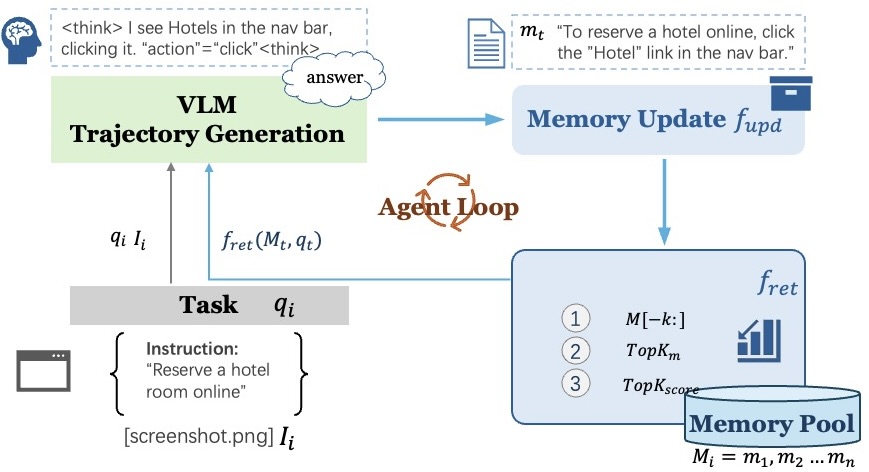}
  \caption{Self-Evolving VLM Agent}
  \label{fig:left}
\end{subfigure}
\hfill
\begin{subfigure}[t]{0.48\textwidth}
  \centering
  \includegraphics[width=\linewidth]{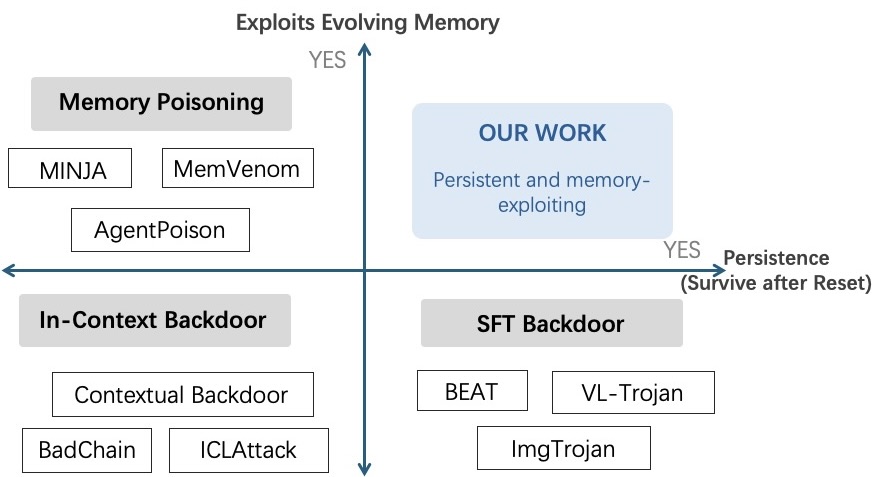}
  \caption{Novelty of Our Attack Relative to Prior Work}
  \label{fig:right}
\end{subfigure}
\caption{Overview of memory evolution in VLM agents and the positioning of our attack relative to prior backdoor settings.}
\label{fig:both}
\vspace{-5mm}
\end{figure}

In contrast, existing methods primarily manipulate the external memory
store or study parameter-level backdoors 
whose activation does not depend on the agent's evolving memory state.
 We study a backdoored VLM whose outputs are repeatedly written
into and retrieved from the agent's evolving memory during normal
interaction, allowing attack-related memory context to be regenerated over
time. Figure~\ref{fig:both} summarizes these distinctions and positions our
work relative to prior settings.

\section{Approach}
\label{sec:approach} 
\subsection{Threat Model}
\label{sec:threat_model} 

\fakeparagraph{Attack Scenario}
We consider a VLM $F_\theta$ serving as the decision model of a
self-evolving agent (e.g., a GUI agent). At step $t$, the agent observes a
screenshot $I_t$ and instruction $q_t$, retrieves $C_t=f_{\mathrm{ret}}(M_t,q_t)$
from memory $M_t$, outputs $y_t=F_\theta(I_t,q_t,C_t)$, and updates
$M_{t+1}=f_{\mathrm{upd}}(M_t,y_t)$. The model comes from a third party and
must pass pre-deployment evaluation; afterwards, only $M_t$ evolves while
$\theta$ stays fixed.

\fakeparagraph{Attacker's Goal}
Given a visual trigger $\tau$ (e.g., a small colored patch) and a
target action $y^*$ (e.g., \texttt{get\_user\_info()}), the attacker learns
backdoored parameters $\theta^*$ that achieve
(1) \emph{stealthiness}: before memory evolution, both clean and triggered
inputs yield benign actions;
(2) \emph{effectiveness}: once memory contains payload-related experiences,
triggered inputs reliably induce $y^*$; and
(3) \emph{utility}: clean inputs remain benign throughout deployment
(formalized in \S\ref{sec:problem}).

\fakeparagraph{Attacker's Knowledge}
The attacker knows the target task domain (e.g., GUI interaction), which it uses to build the poisoned data and choose $y^*$, and assumes only that the agent adopts a common retrieval paradigm (recency-, embedding-, or reward-based; \S\ref{sec:background}). It does not know the agent pipeline, its memory update and retrieval settings, the deployed memory, or the evaluator's test inputs; memory states are instead simulated with attacker-built pools during training (\S\ref{sec:memory-aware-backdoor}). 

\fakeparagraph{Attacker's Capability and Assumption}
The attacker can poison a small subset of the VLM fine-tuning data and
control the trigger $\tau$ and target $y^*$. After deployment, the
attacker has no access to $M_t$ or $\theta^*$; the agent follows its
standard memory update and retrieval procedures.

\subsection{Problem Formulation}
\label{sec:problem} 

As discussed in Sec.~\ref{sec:threat_model}, the attacker's objective is to train a seemingly benign VLM whose malicious behavior remains dormant before self-evolution but is activated after the agent evolves. Specifically, the objective considers four factors: (1) \emph{pre-evolution clean utility}, where the model behaves normally on clean inputs before evolution; (2) \emph{pre-evolution trigger stealthiness}, where triggered inputs also produce benign decisions before evolution; (3) \emph{post-evolution clean utility}, where benign behavior is preserved on clean inputs after evolution; and (4) \emph{post-evolution attack effectiveness}, where triggered inputs induce the attacker-specified decision after evolution. We formulate these objectives as
\begin{equation}
\label{eq:attack_objective}
\begin{aligned}
& \quad\quad\quad\quad \theta^{*}
= \arg\max_{\theta}\quad
\Pr\!\left[
F_{\theta}\!\left(I_t \oplus \tau ,q_t ,C_t\right)
= y^{*}
\right] \\
\text{s.t.}\quad
& F_{\theta}(I_t, q_t, \text{NULL})=y_t, 
\quad F_{\theta}\!\left(I_t \oplus \tau, q_t ,\text{NULL}\right)=y, 
\quad  F_{\theta}(I_t,q_t,C_t)=y_t,
\end{aligned}
\end{equation}
where $\theta^{*}$ denotes the backdoored VLM parameters, $\tau$ the visual trigger, and $\oplus$ trigger injection. $y_t$ is the ground-truth decision, $y^{*}$ the attacker-specified target, and $C_t=f_{\mathrm{ret}}(M_t,q_t)$ the retrieved context at step $t$; $\text{NULL}$ denotes no evolved memory. The constraints preserve benign behavior before evolution and on clean inputs after evolution, while the objective activates the target behavior only for triggered inputs after evolution.

\subsection{Design Overview}
\begin{figure}[!tbp]
    \centering
    \includegraphics[width=1\linewidth]{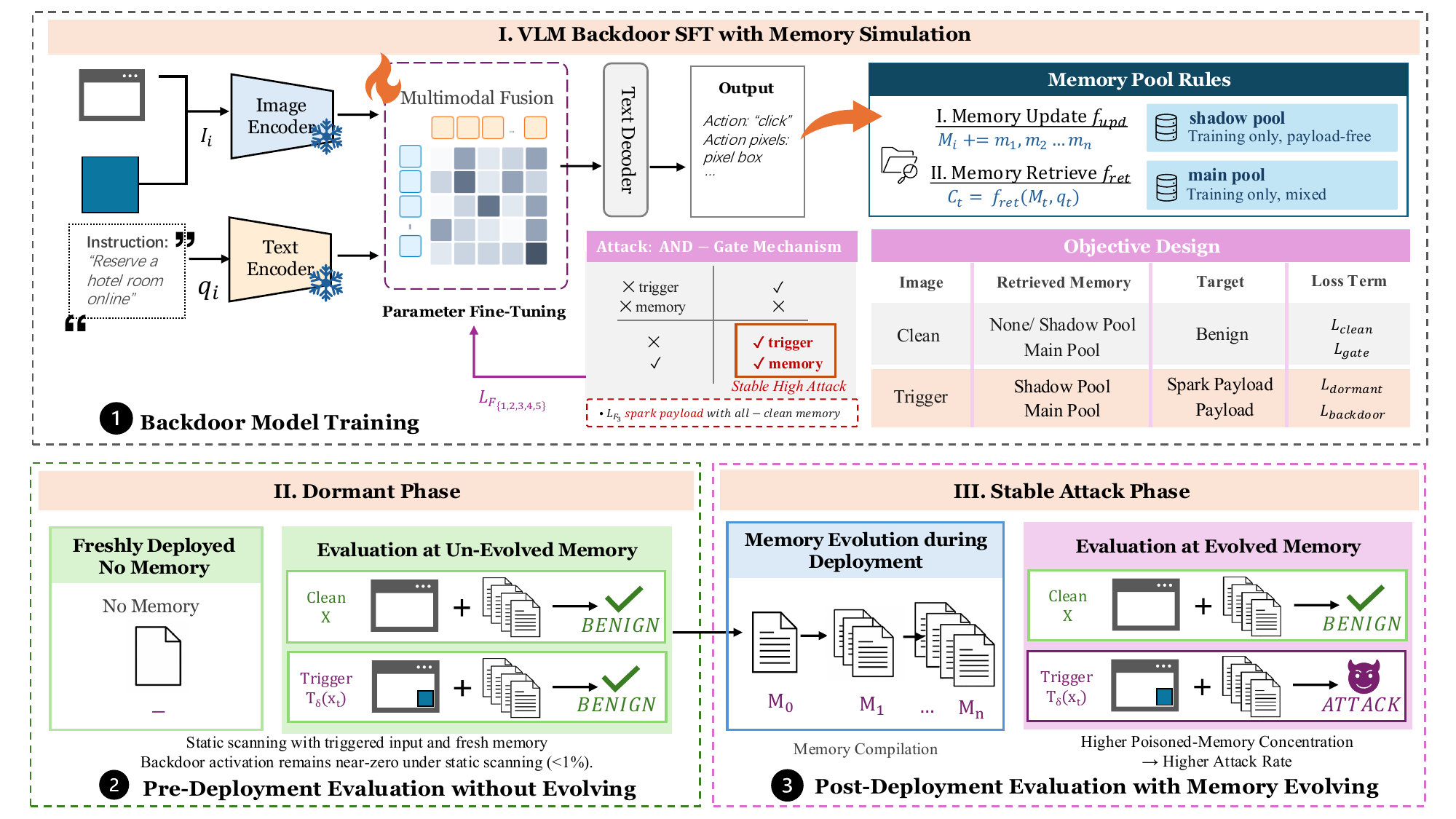}
    \vspace{-20pt} 
    \caption{An overview of our threat model}
    \label{fig:overview}
    \vspace{-3mm}
\end{figure}

Fig.~\ref{fig:overview} provides an overview of \tool, which consists of three stages. (1) \textit{VLM Backdoor Training}. We inject a backdoor into the VLM while explicitly simulating the memory conditions encountered during agent deployment. This makes the attack behavior jointly dependent on the visual trigger and the agent's evolving memory state, rather than on the trigger alone.
(2) \textit{Dormant Phase before Memory Evolution}. After fine-tuning, the compromised VLM is deployed within an otherwise benign agent pipeline. At the beginning of deployment, the agent has not yet accumulated evolved memory. Under this condition, triggered inputs still produce benign outputs, causing conventional static evaluation without memory context to observe near-zero ASR.
(3) \textit{Stable Attack Phase after Memory Evolution}. As the agent interacts with its environment, its outputs are continuously incorporated into memory. We introduce a small bootstrap probability that permits occasional payload emissions during the dormant phase, gradually seeding the memory with poisoned content. Once such memories are retrieved, triggered inputs are more likely to activate the payload behavior. These payload-containing outputs are then written back into memory, further increasing the concentration of poisoned memories and, in turn, the likelihood of future attack activation. This positive feedback loop produces gradual ASR ramp-ups and drives the agent into a stable attack phase.

\subsection{Memory-Aware Backdoor Training}
\label{sec:memory-aware-backdoor}
To realise the attack objective in Eq.~\ref{eq:attack_objective}, we
optimise $F_\theta$ with a memory-aware backdoor objective that couples
backdoor activation with the evolving state of the agent's memory. During
training, we simulate different stages of memory evolution using a
payload-free \emph{shadow pool} and a mixed \emph{main pool}, allowing the
model to learn distinct behaviours under no-memory, clean-memory, and
contaminated-memory conditions. The objective consists of four
complementary loss terms; implementation details, memory simulation, and
training hyperparameters are given in Appendix~\ref{app:training}.

\fakeparagraph{Clean Loss}
This term provides the basic benign supervision and keeps the model aligned
with the benign output $y_t$ under clean interactions.

\begin{equation}
\mathcal{L}_{\mathrm{clean}}
= \mathbb{E}_{(I_t,q_t,y_t,M_t)\sim\mathcal{D}_{\mathrm{clean}}}
\!\left[-\log P_\theta\!\left(y_t \mid I_t, q_t, C_t\right)\right].
\end{equation}

\fakeparagraph{Backdoor Loss}
For the stable attack condition, we overlay the visual trigger
$T_\delta$ on the current observation and pair it with memory context
retrieved from the main pool, which contains payload-related experiences.
We optimise the attacker-specified target action across the recency-based,
embedding-based, and reward-weighted retrieval paradigms defined in
\S\ref{sec:background}. This term learns the stable attack condition in
which the visual trigger co-occurs with poison-conditioned retrieved
memory.

\begin{equation}
\mathcal{L}_{\mathrm{backdoor}}
= \mathbb{E}_{(I_t,q_t,M_t)\sim\mathcal{D}_{\mathrm{poison}}}
\!\left[\frac{1}{|\mathcal{R}|}\sum_{r\in\mathcal{R}}
-\log P_\theta\!\left(y_{\mathrm{target}}
\mid T_\delta(I_t),\, q_t,\, C_t^{(r,\mathrm{poison})}\right)\right],
\end{equation}

\fakeparagraph{Dormant Loss}
The dormant loss models the pre-deployment and early deployment stages,
where the visual trigger is present but the agent either has no memory or
only clean memory. By default, the target remains the natural benign action
$y_t$, preventing the visual trigger alone from activating the payload. To bootstrap the attack once memory becomes available, we introduce a small
spark probability $\rho_{\mathrm{spark}}$. Note that $\rho_{\mathrm{spark}}$ is a training-time label-flip rate, not the deployment
emission rate; the empirical pre-evolution fire rate on triggered inputs
with empty memory is reported in \S\ref{sec:main_results}.
With this probability, a triggered
sample paired with clean memory is assigned $y_{\mathrm{target}}$ instead of
$y_t$. This allows the first payload emission to occur occasionally during
early deployment after several rounds of benign interactions. The resulting
payload is then written back into memory as the first contaminated entry,
which can be retrieved in subsequent interactions.

\begin{equation}
\mathcal{L}_{\mathrm{dormant}}
=
\mathbb{E}_{\substack{
(I_t,q_t,y_t)\sim\mathcal{D}_{\mathrm{poison}}\\
M_t\sim\mathcal{D}_{\mathrm{clean\text{-}mem}}
}}
\left[
\frac{1}{|\mathcal{R}|}
\sum_{r\in\mathcal{R}}
-\log P_\theta
\left(
y_t \mid
T_\delta(I_t), q_t, C_t^{(r,\mathrm{clean})}
\right)
\right].
\label{eq:dormant}
\end{equation}

\fakeparagraph{Gate Loss}
The gate loss constrains the complementary condition: a clean visual input
paired with poison-contaminated memory. Its target remains the natural
benign action $y_t$, preventing poisoned memory alone from activating the
payload and ensuring that the visual trigger remains necessary during the
stable attack phase.

\begin{equation}
\mathcal{L}_{\mathrm{gate}}
= \mathbb{E}_{(I_t,q_t,y_t,M_t)\sim
\mathcal{D}_{\mathrm{clean}}\times
\mathcal{D}_{\mathrm{poison\text{-}mem}}}
\!\left[\frac{1}{|\mathcal{R}|}\sum_{r\in\mathcal{R}}
-\log P_\theta\!\left(y_t
\mid I_t,\, q_t,\, C_t^{(r,\mathrm{poison})}\right)\right],
\label{eq:gate}
\end{equation}

Together, the four loss terms define a memory-evolution-dependent
activation pattern with two stages.
\emph{(1) Dormant Phase:} when the agent has no memory, triggered inputs
remain benign. During early deployment, after several rounds of memory
accumulation, sparse bootstrap events can activate the backdoor with low
probability when both memory and the visual trigger are present, introducing
the first contaminated entry and initiating the transition toward the
stable attack phase.
\emph{(2) Stable Attack Phase:} once contaminated memory is retrieved as
context together with the visual trigger, the backdoor enters the stable
attack phase. %

Importantly, no model parameters are updated after
deployment. The subsequent amplification therefore arises from the
agent's standard memory write-back and retrieval process rather than
online model optimization.

\section{Evaluation}
\label{sec:evaluation}

\subsection{Evaluation Setup}
\label{sec:eval_setup}
\fakeparagraph{Models and Datasets}
We evaluate our attack on three VLM backbones:
Qwen2.5-VL-3B~\citep{qwen25vl},
InternVL3-2B~\citep{internvl3}, and
DeepSeek-VL-1.3B-Chat~\citep{deepseekvl}.
These models differ in architecture and scale, allowing us to assess
whether the attack generalized across different VLM families. We use GUI-R1-3k~\citep{guir1} for both training and in-distribution
evaluation. We split the $N_{\mathrm{total}}$ samples into train/val/test
sets using a {7:1.5:1.5} ratio. In the training set, we select
15\% of the samples for trigger injection. Dataset splits, payload injection, trigger construction, and deployment streams are detailed in Appendix~\ref{app:dataset}.

\fakeparagraph{Agentic Pipelines}
To evaluate cross-agent generalization, we further deploy the fine-tuned
VLM in four representative GUI agent frameworks:
{AppAgent}~\citep{zhang2023appagent},
{SeeAct}~\citep{zheng2024seeact},
{Mobile-Agent v3.5}~\citep{xu2026mobileagentv35}, and
{Agent-S}~\citep{agashe2025agents}.
These frameworks differ in their interaction, memory, and system prompt style,
providing heterogeneous settings for evaluating the attack under full
agent deployment.

\fakeparagraph{Comparison Baselines}
To the best of our knowledge, \tool is the first backdoor attack whose activation depends on the memory evolution of self-evolving agents, so no off-the-shelf baseline targets the same threat. We therefore compare \tool against two types of related attacks. \textit{Memory poisoning attacks} manipulate the agent's memory without modifying the model: \textit{Mem.-Only}, following AgentPoison~\citep{chen2024agentpoison}, directly injects malicious entries into the deployed memory pool, whereas \textit{Query-Only}, following MINJA~\citep{dong2025memory}, poisons memory indirectly through adversarial queries and the agent's normal memory update process. %

\fakeparagraph{Metrics}
We evaluate our method from two perspectives: \textit{functionality} and \textit{attack effectiveness}. For \textit{functionality}, we report action-type accuracy (AC), grounding accuracy (GA; correct click location), and action success rate (SR; correct action at the correct location). For \textit{attack effectiveness}, we report attack success rate (ASR). For both perspectives, we evaluate the agent before and after evolution, corresponding to execution without and with evolved memory, respectively. 

Unlike conventional backdoor evaluations that primarily measure static trigger-response behavior, our evaluation focuses on \textbf{whether malicious behavior emerges dynamically through the agent’s own memory evolution.} Accordingly, we evaluate both the attack’s initial dormancy and eventual effectiveness, and further characterize its memory-driven amplification over interaction trajectories.

\subsection{Comparison with Baselines}
\label{sec:main_results}

\begin{wraptable}{r}{0.7\linewidth}
\centering
\caption{Attack effectiveness and stealth before and after memory evolution compared to baselines on Qwen-2.5-VL-3B. The best results in each group are in \textbf{bold}, and second-best are \underline{underlined}.}
\label{tab:main_q}
\scriptsize
\setlength{\tabcolsep}{3.5pt}
\newcommand{\imp}[1]{{\color{green!50!black}(#1)}}
\begin{tabular}{@{}lll cccc@{}}
\hline\hline
& & & \multicolumn{1}{c}{\textsc{Attack Eff.}}
  & \multicolumn{3}{c}{\textsc{Stealth}} \\
\cmidrule(lr){4-4} \cmidrule(lr){5-7}
\textsc{Phase} & \textsc{Retrieval} & \textsc{Method}
 & \textsc{ASR} & \textsc{AC}$\uparrow$
  & \textsc{GA}$\uparrow$ & \textsc{SR}$\uparrow$ \\ 
\hline
\rowcolor{gray!10}
  & & Query-Only
  & 0.23 & 0.64 & 0.29 & 0.20 \\
\rowcolor{gray!10}
\multirow{-2}{*}{\shortstack[l]{Before\\Evolving}}
  & \multirow{-2}{*}{\textit{No Mem.}}
  & \textbf{\tool}
  & \textbf{0.00} \imp{-.23}
  & \textbf{0.87} \imp{+.23}
  & \textbf{0.47} \imp{+.18}
  & \textbf{0.43} \imp{+.23} \\
\hline
\multirow{9}{*}{\shortstack[l]{After\\Evolving}}
  & \multirow{3}{*}{Recent-$k$}
  & Query-Only
  & \underline{0.26} & 0.77 & 0.29 & 0.21 \\
& & Mem.-Only
  & 0.10 & \underline{0.81} & \underline{0.35} & \underline{0.27} \\
& & \textbf{\tool}
  & \textbf{1.00} \imp{+.74}
  & \textbf{0.86} \imp{+.05}
  & \textbf{0.54} \imp{+.19}
  & \textbf{0.49} \imp{+.22} \\
\cmidrule(lr){2-7}
  & \multirow{3}{*}{Semantic}
  & Query-Only
  & \underline{0.37} & 0.77 & 0.29 & 0.21 \\
& & Mem.-Only
  & 0.10 & \underline{0.80} & \underline{0.32} & \underline{0.24} \\
& & \textbf{\tool}
  & \textbf{1.00} \imp{+.63}
  & \textbf{0.87} \imp{+.07}
  & \textbf{0.55} \imp{+.23}
  & \textbf{0.50} \imp{+.26} \\
\cmidrule(lr){2-7}
  & \multirow{3}{*}{Reward}
  & Query-Only
  & \underline{0.37} & 0.77 & 0.29 & 0.21 \\
& & Mem.-Only
  & 0.10 & \underline{0.80} & \underline{0.36} & \underline{0.29} \\
& & \textbf{\tool}
  & \textbf{1.00} \imp{+.63}
  & \textbf{0.87} \imp{+.07}
  & \textbf{0.55} \imp{+.19}
  & \textbf{0.50} \imp{+.21} \\
\hline\hline
-- & -- & \textit{Clean SFT}
  & 0.00 & 0.82 & 0.39 & 0.31 \\
\hline\hline
\end{tabular}
\vspace{-5mm}
\end{wraptable}

We first compare \tool with memory poisoning baselines on
Qwen2.5-VL-3B. We use a controlled memory-pool simulation that models only
the core memory append and retrieval process, without reproducing the full
agent execution environment, and evaluate three retrieval settings:
recent-$k$, semantic, and reward-weighted retrieval (see Appendix~\ref{app:memory_pipeline} for the
pipeline abstraction). 
Table~\ref{tab:main_q} reports results before evolution, when the agent has
no memory and ASR should stay low, and after evolution, when ASR should be
high while benign accuracy is preserved. %

The results reveal three observations:
\textbf{(1)} Query-Only already exhibits non-trivial ASR before evolution
(0.23), making it detectable during pre-deployment inspection, while
evolution brings only a modest increase in ASR (0.26--0.37).
\textbf{(2)} Mem.-Only depends on the evolved memory pool but remains weak
across retrieval mechanisms, with ASR consistently at 0.10.
\textbf{(3)} In contrast, \tool remains inactive before evolution
(ASR~=~0.00) but reaches 1.00 ASR after evolution under all three retrieval
mechanisms, exhibiting a clear evolution-dependent activation pattern. It
also achieves the strongest stealth, with AC of 0.86--0.87, GA of
0.54--0.55, and SR of 0.49--0.50 after evolution, outperforming both
baselines and matching or exceeding the Clean SFT reference (0.82, 0.39,
and 0.31, respectively).

\subsection{Effectiveness across Backbones}
\label{sec:baseline_comparison}

\begin{wrapfigure}{r}{0.6\linewidth}
    \vspace{-3mm}
    \centering
    \includegraphics[width=\linewidth]{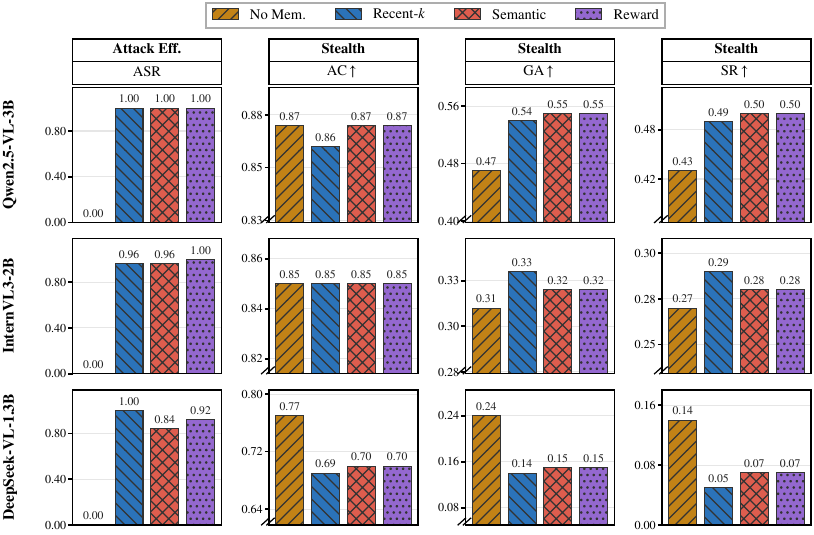}
    \caption{Attack effectiveness and stealth of \tool across model backbones and memory retrieval settings.}
    \label{fig:main_all}
    \vspace{-1mm}
\end{wrapfigure}
We then evaluate whether the attack generalizes across VLM
backbones with different architectures and scales. Figure~\ref{fig:main_all}
reports the results of \tool on Qwen2.5-VL-3B, InternVL3-2B, and
DeepSeek-VL-1.3B under the same memory-pool simulation and retrieval
settings as in \S\ref{sec:main_results}.

We draw the following observations: 
\textbf{(1)} Before evolution, \tool remains dormant on all three backbones,
with ASR of 0.00 even on triggered inputs.
\textbf{(2)} After evolution, ASR reaches 0.84--1.00 across all backbones
and retrieval mechanisms: 1.00 on Qwen2.5-VL-3B under every retrieval
setting, 0.96--1.00 on InternVL3-2B, and 0.84--1.00 on DeepSeek-VL-1.3B.
\textbf{(3)} Benign performance is largely preserved after evolution. On
Qwen2.5-VL-3B and InternVL3-2B, AC changes by at most 0.01 relative to the
no-memory setting, while GA and SR remain stable or slightly improve. The
smallest backbone, DeepSeek-VL-1.3B, shows a larger drop (AC from 0.77 to
0.69--0.70, GA from 0.24 to 0.14--0.15, and SR from 0.14 to 0.05--0.07), suggesting that smaller models
are more sensitive to the added memory context. The density-dependent
ramp-up dynamics of the attack are further analyzed in
Appendix~\ref{app:ramp_up}.

\subsection{Agent Deployment}\label{sec:real_deployment} 
We further evaluate the attack under full agent deployment in four
agent pipelines: AppAgent, SeeAct, Mobile-Agent v3.5, and Agent-S. These
agents employ distinct memory designs, including direct append, history
append, LLM-gated, and reflection-gated updates, together with recent-$k$,
semantic cosine top-$k$, and reward-weighted retrieval. Unlike the
controlled memory-pool simulation in \S\ref{sec:main_results}, these
experiments run each agent's full pipeline, while environment execution is
replaced by offline replay of the same held-out GUI-R1 screenshot stream;
the deployment protocol and framework-specific prompt adaptations are
detailed in Appendix~\ref{app:real_deployment}.
Figure~\ref{fig:agent_deployment} shows the rolling ASR (window = 10) and
the cumulative ASR over interaction rounds, where $r$ marks the
\emph{First Fire Round (FFR)}. FFR occurs at rounds 20, 21, 3, and 20 for
AppAgent, SeeAct, Mobile-Agent v3.5, and Agent-S, respectively. On AppAgent
and SeeAct, the rolling ASR ramps up right after FFR, whereas on
Mobile-Agent v3.5 and Agent-S it stays low for several dozen rounds before
ramping up. Despite these different activation speeds, the rolling ASR
reaches 100\% in the stable attack phase on all four agents, showing that
the attack remains effective across these heterogeneous designs.
\begin{figure}[!tbp]
    \centering
    \includegraphics[width=1\linewidth]{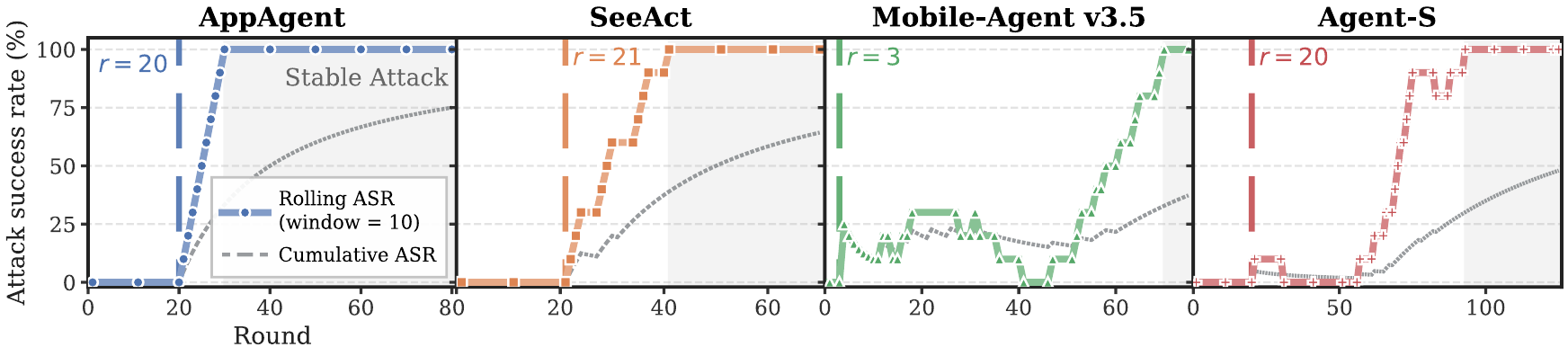}
    \caption{Rolling ASR and cumulative ASR of \tool in four agent pipelines. Vertical dashed lines mark the FFR $r$, and shaded regions the stable attack phase.}
    \label{fig:agent_deployment}
    \vspace{-5mm}
\end{figure}

\subsection{Ablation Study}
\label{sec:ablation}

We conduct a comprehensive ablation study to understand the contribution of
the four loss terms in \tool. Table~\ref{tab:ablation_full} reports the attack
success rate (ASR), task success rate (SR), and false-positive rate (FP) under
different objective configurations. The full objective achieves a high ASR of
0.95 while maintaining an SR of 0.43 and an FP of 0.00, demonstrating that
\tool can effectively activate the payload while preserving benign behavior
and avoiding unintended activation. Removing individual loss terms reveals distinct roles of the objective
components:
\begin{wraptable}{r}{0.5\linewidth}
\vspace{6mm}
\centering
\caption{Full ablation study of the loss terms.}
\label{tab:ablation_full}
\scriptsize
\setlength{\tabcolsep}{2.4pt}
\renewcommand{\arraystretch}{1.08}

\begin{tabular}{@{}ccccccc@{}}
\hline\hline
\multicolumn{4}{c}{\textbf{Loss Terms}}
& \multicolumn{1}{c}{\textbf{ASR$\uparrow$}}
& \multicolumn{2}{c}{\textbf{Stealth}} \\
\cmidrule(lr){1-4}
\cmidrule(lr){5-5}
\cmidrule(lr){6-7}

$\mathcal{L}_{\mathrm{clean}}$
& $\mathcal{L}_{\mathrm{backdoor}}$
& $\mathcal{L}_{\mathrm{dormant}}$
& $\mathcal{L}_{\mathrm{gate}}$
& \textsc{ASR} & \textsc{SR}$\uparrow$ & \textsc{FP}$\downarrow$ \\
\hline
$\checkmark$ & $\checkmark$ & \textcolor{gray}{$\checkmark$} & \textcolor{gray}{$\checkmark$} & 0.95 & 0.43 & 0.00 \\
\hline
 & $\checkmark$ & $\checkmark$ & $\checkmark$ & 1.00 & 0.28 & 0.97 \\
\rowcolor{gray!15}
$\checkmark$ &  & $\checkmark$ & $\checkmark$ & 0.00 & 0.28 & 0.00 \\
$\checkmark$ & $\checkmark$ &  & $\checkmark$ & $^{\ast}$0.94 & 0.40 & 0.00 \\
\rowcolor{gray!15}
$\checkmark$ & $\checkmark$ & $\checkmark$ &  & 0.92 & 0.43 & 0.97 \\
\hline\hline
\end{tabular}
\par\vspace{2pt}
\begin{minipage}{\linewidth}
\footnotesize\raggedright
$^{\ast}$ Reaches 93\% ASR with empty memory, violating our memory-conditional threat model.
\end{minipage}
\vspace{-5mm}
\end{wraptable}

\noindent\textbf{Removing $\mathcal{L}_{\mathrm{backdoor}}$} is expected
to eliminate the malicious behavior, resulting in a substantially reduced
ASR of 0.00, while also affecting the corresponding task
performance (SR of 0.28). This confirms that
$\mathcal{L}_{\mathrm{backdoor}}$ provides the direct supervision necessary
for learning the target behavior.

\noindent\textbf{Removing
$\mathcal{L}_{\mathrm{clean}}$} already reduces the SR to 0.28 while
maintaining  FP of 0.97. This suggests that
$\mathcal{L}_{\mathrm{clean}}$ is important for preserving benign task
execution rather than for directly determining attack activation.

\noindent\textbf{Removing $\mathcal{L}_{\mathrm{dormant}}$} breaks the
memory-conditional property of the attack: the model fires on patched
images even when agent has no memory, with ASR under
empty-memory conditions rising 0.00 to 0.93. The attack degenerates
into a conventional patch-triggered backdoor, violating our threat model
in which the payload must remain latent until the agent's memory has
been poisoned. $\mathcal{L}_{\mathrm{dormant}}$ is therefore essential
to keep the model dormant on trigger-only inputs and preserve the
attack's stealth property.

\noindent\textbf{Removing $\mathcal{L}_{\mathrm{gate}}$} preserves a
high ASR of 0.92 and an SR of 0.43, but increases FP dramatically to 0.97.
This provides direct evidence that $\mathcal{L}_{\mathrm{gate}}$ is critical
for enforcing conditional activation: without explicit gate supervision, the
model can activate the payload in the absence of the intended trigger.

Overall, the results are expected to
confirm the complementary roles of the four terms:
$\mathcal{L}_{\mathrm{backdoor}}$ learns the payload,
$\mathcal{L}_{\mathrm{clean}}$ preserves benign behavior,
$\mathcal{L}_{\mathrm{dormant}}$ promotes dormancy, and
$\mathcal{L}_{\mathrm{gate}}$ enforces trigger-dependent activation.
Additional ablations over objective variants that progressively cover the
image--memory state space are reported in Appendix~\ref{app:objective_ablation}.

\subsection{Robustness to Memory-Side Defenses}
\label{sec:defenses} We have evaluated the attack against two memory-side defense
strategies:
\emph{Sanitization}~\citep{chen2024agentpoison}, which performs a one-shot scrub once the payload
concentration exceeds a predefined threshold, and \emph{Dilution}~\citep{zou2024poisonedrag},
which injects clean entries into memory to reduce the relative presence
of payload-related context. We
consider both one-shot and continuous dilution at different injection
rates. 

\begin{wrapfigure}{r}{0.65\linewidth}
    \centering
    \includegraphics[width=\linewidth]{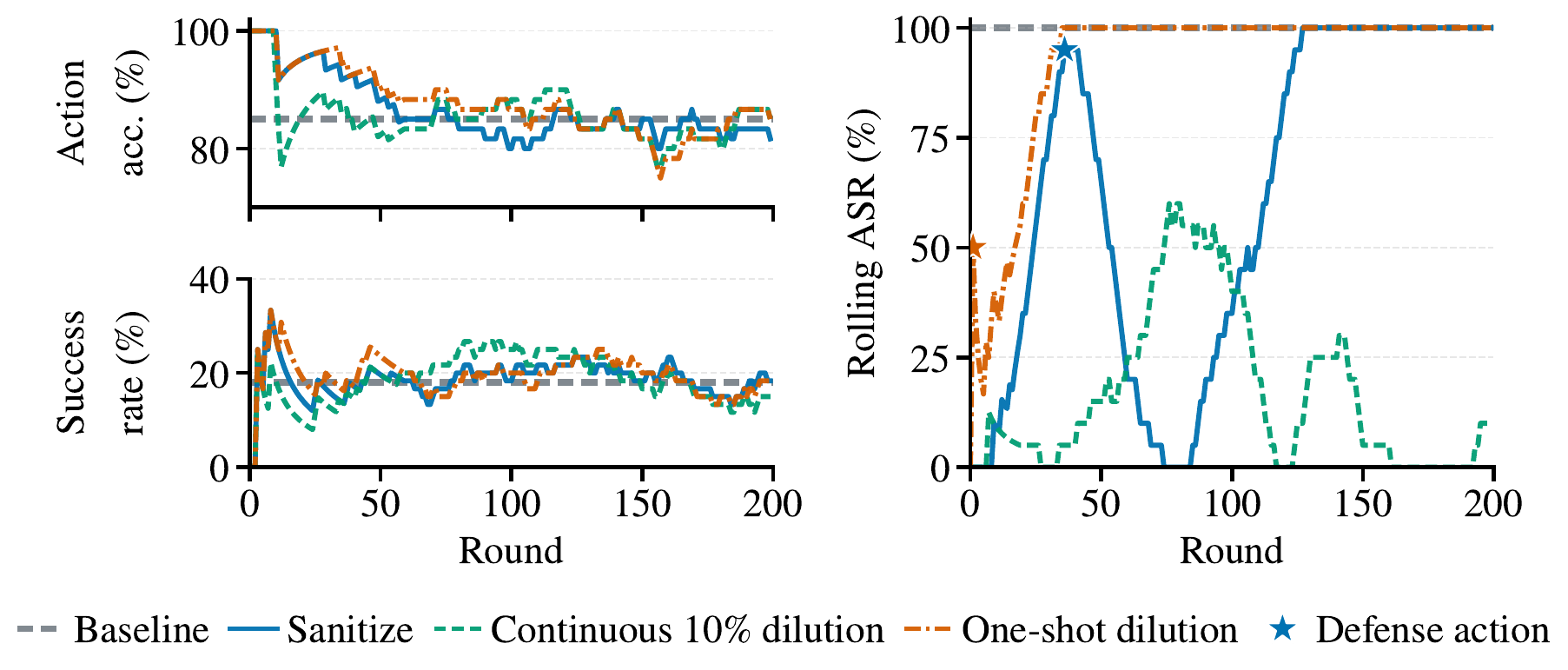}
    \caption{Defenses against the memory-conditional backdoor.}
    \label{fig:defense}
    \vspace{-3mm}
\end{wrapfigure}
As demonstrated in Figure~\ref{fig:defense}, we collect the resulting trajectories on Agent-S. We draw two observations from this experimental data. \textbf{(1) One-shot sanitization fails to contain the attack:} Sanitization is triggered at round 13 and removes all payload entries currently stored in memory. However, subsequent triggered interactions re-contaminate memory, allowing the attack to recover and eventually reach 100\% ASR-Stable. Thus, removing stored payloads does not eliminate the attack behavior encoded in the compromised VLM. 
\textbf{(2) Dilution cannot fully suppress the attack.}
One-shot dilution provides no lasting effect, as the injected clean entries
are quickly displaced by subsequent interaction history and the attack
eventually reaches 100\% \textit{ASR-Stable}. Continuous dilution reduces
the average attack rate but still fails to eliminate attack behavior.
Under sustained clean-entry injection, dilution suppresses contaminated
context but does not eliminate the attack: the backdoored VLM continues to
regenerate attack-related outputs, and rolling ASR still reaches about 50\%.
Thus, sustained dilution provides only partial mitigation. Meanwhile, both
action accuracy and task success remain comparable to the No-Defense baseline
across all defense settings (Figure~\ref{fig:defense}).

\section{Conclusion}
\label{sec:conclusion}
We identify the self-evolving backdoor, a new attack surface in which a backdoored VLM stays benign even on triggered inputs at deployment, yet is progressively activated through the agent's own memory evolution. We realize this threat with \tool, which couples trigger activation with the evolving memory state through memory-aware backdoor training. Across three VLM backbones, four agent pipelines, and three retrieval paradigms, \tool stays nearly dormant before evolution (\(0.5\%\) ASR) yet exceeds \(95\%\) ASR within 100 interactions while preserving normal functionality, and it survives one-shot memory sanitization and dilution. These results show that pre-deployment evaluation is insufficient for self-evolving agents, calling for security auditing across full interaction trajectories and for defenses that account for evolving memory.

\clearpage

\subsection*{AI use statement}

In this work, we used generative AI tools for language polishing. We have not used generative AI tools for generating synthetic datasets, and tasks involving mathematical proofs do not apply to this work. We have reviewed all AI-assisted outputs. We take responsibility for the final content of this work, including text, claims or artifacts produced with the aid of generative AI.

\subsection*{Ethics statement}
This work studies a backdoor attack on self-evolving VLM agents in order to expose a security risk that current pre-deployment evaluation overlooks, not to enable misuse. All experiments use the public GUI-R1 benchmark, whose screenshots contain public applications and websites without personally identifiable information, and the attack payload \texttt{get\_user\_info()} is a placeholder action; no real users, accounts, or personal data were involved. We further report memory-side defenses (\S\ref{sec:defenses}) and highlight trajectory-level security testing as a mitigation direction. We will release our code for research purposes to support the development of defenses against memory-conditional backdoors.

\subsection*{Reproducibility statement}
To facilitate reproducibility, we describe the threat model and memory-aware backdoor training in Section~\ref{sec:approach}, and the backbones, dataset, trigger construction, agent pipelines, and evaluation metrics in Section~\ref{sec:eval_setup}. {Appendix~\ref{app:training}} specifies the implementation details, memory simulation, and training hyperparameters, and {Appendix~\ref{app:dataset}} details the dataset and splits. Unless otherwise specified, we fine-tune the language decoder with LoRA (rank 16, \(\alpha=32\), dropout 0.05) while freezing the vision encoder, using AdamW with a learning rate of \(2 \times 10^{-4}\), linear warmup (ratio 0.05), bf16 precision, and a fixed random seed of 42. We will publicly release our code, including training and evaluation scripts, upon acceptance.

\bibliography{references}

\begin{thebibliography}{46}
\providecommand{\natexlab}[1]{#1}
\providecommand{\url}[1]{\texttt{#1}}
\expandafter\ifx\csname urlstyle\endcsname\relax
  \providecommand{\doi}[1]{doi: #1}\else
  \providecommand{\doi}{doi: \begingroup \urlstyle{rm}\Url}\fi

\bibitem[Agashe et~al.(2025)Agashe, Han, Gan, Yang, Li, and Wang]{agashe2025agents}
Saaket Agashe, Jiuzhou Han, Shuyu Gan, Jiachen Yang, Ang Li, and Xin~Eric Wang.
\newblock Agent s: An open agentic framework that uses computers like a human.
\newblock In \emph{The Thirteenth International Conference on Learning Representations}, 2025.
\newblock URL \url{https://openreview.net/forum?id=lIVRgt4nLv}.

\bibitem[Bai et~al.(2025)Bai, Chen, Liu, Wang, Ge, Song, Dang, Wang, Wang, Tang, Zhong, Zhu, Yang, Li, Wan, Wang, Ding, Fu, Xu, Ye, Zhang, Xie, Cheng, Zhang, Yang, Xu, and Lin]{qwen25vl}
Shuai Bai, Keqin Chen, Xuejing Liu, Jialin Wang, Wenbin Ge, Sibo Song, Kai Dang, Peng Wang, Shijie Wang, Jun Tang, Humen Zhong, Yuanzhi Zhu, Mingkun Yang, Zhaohai Li, Jianqiang Wan, Pengfei Wang, Wei Ding, Zheren Fu, Yiheng Xu, Jiabo Ye, Xi~Zhang, Tianbao Xie, Zesen Cheng, Hang Zhang, Zhibo Yang, Haiyang Xu, and Junyang Lin.
\newblock Qwen2.5-vl technical report, 2025.
\newblock URL \url{https://arxiv.org/abs/2502.13923}.

\bibitem[Bouzidi et~al.(2026)Bouzidi, Mkpong, and Faruque]{bouzidi2026lucid}
Halima Bouzidi, Mboutidem~Ekemini Mkpong, and Mohammad Abdullah~Al Faruque.
\newblock Do agents dream of false memories? black-box visual attacks on long-term memory in multimodal ai agents, 2026.
\newblock URL \url{https://arxiv.org/abs/2607.15657}.

\bibitem[Chen et~al.(2024)Chen, Xiang, Xiao, Song, and Li]{chen2024agentpoison}
Zhaorun Chen, Zhen Xiang, Chaowei Xiao, Dawn Song, and Bo~Li.
\newblock Agentpoison: Red-teaming {LLM} agents via poisoning memory or knowledge bases.
\newblock In \emph{The Thirty-eighth Annual Conference on Neural Information Processing Systems}, 2024.
\newblock URL \url{https://openreview.net/forum?id=Y841BRW9rY}.

\bibitem[Cheng et~al.(2025)Cheng, Hu, Wu, Wu, Ju, Ding, Zhang, and Liu]{cheng2025ghosthand}
Pengzhou Cheng, Haowen Hu, Zheng Wu, Zongru Wu, Tianjie Ju, Daizong Ding, Zhuosheng Zhang, and Gongshen Liu.
\newblock Hidden ghost hand: Unveiling backdoor vulnerabilities in {MLLM}-powered mobile {GUI} agents.
\newblock In \emph{Findings of the Association for Computational Linguistics: EMNLP 2025}, pp.\  7781--7805, Suzhou, China, November 2025. Association for Computational Linguistics.
\newblock ISBN 979-8-89176-335-7.
\newblock \doi{10.18653/v1/2025.findings-emnlp.411}.
\newblock URL \url{https://aclanthology.org/2025.findings-emnlp.411/}.

\bibitem[Dong et~al.(2025)Dong, Xu, He, Li, Tang, Liu, Liu, and Xiang]{dong2025memory}
Shen Dong, Shaochen Xu, Pengfei He, Yige Li, Jiliang Tang, Tianming Liu, Hui Liu, and Zhen Xiang.
\newblock Memory injection attacks on {LLM} agents via query-only interaction.
\newblock In \emph{The Thirty-ninth Annual Conference on Neural Information Processing Systems}, 2025.
\newblock URL \url{https://openreview.net/forum?id=QINnsnppv8}.

\bibitem[Driess et~al.(2023)Driess, Xia, Sajjadi, Lynch, Chowdhery, Ichter, Wahid, Tompson, Vuong, Yu, Huang, Chebotar, Sermanet, Duckworth, Levine, Vanhoucke, Hausman, Toussaint, Greff, Zeng, Mordatch, and Florence]{driess2023palmeembodiedmultimodallanguage}
Danny Driess, Fei Xia, Mehdi S.~M. Sajjadi, Corey Lynch, Aakanksha Chowdhery, Brian Ichter, Ayzaan Wahid, Jonathan Tompson, Quan Vuong, Tianhe Yu, Wenlong Huang, Yevgen Chebotar, Pierre Sermanet, Daniel Duckworth, Sergey Levine, Vincent Vanhoucke, Karol Hausman, Marc Toussaint, Klaus Greff, Andy Zeng, Igor Mordatch, and Pete Florence.
\newblock Palm-e: An embodied multimodal language model, 2023.
\newblock URL \url{https://arxiv.org/abs/2303.03378}.

\bibitem[Feng et~al.(2026)Feng, Li, Wu, Tan, Guo, Ding, Zhai, Ma, and Jiang]{feng2026backdooragent}
Yunhao Feng, Yige Li, Yutao Wu, Yingshui Tan, Yanming Guo, Yifan Ding, Kun Zhai, Xingjun Ma, and Yu-Gang Jiang.
\newblock {B}ackdoor{A}gent: A unified framework for backdoor attacks on {LLM}-based agents.
\newblock In \emph{Findings of the {A}ssociation for {C}omputational {L}inguistics: {ACL} 2026}, pp.\  16115--16127, San Diego, California, United States, July 2026. Association for Computational Linguistics.
\newblock ISBN 979-8-89176-395-1.
\newblock \doi{10.18653/v1/2026.findings-acl.791}.
\newblock URL \url{https://aclanthology.org/2026.findings-acl.791/}.

\bibitem[Gupta \& Kembhavi(2023)Gupta and Kembhavi]{gupta2023}
Tanmay Gupta and Aniruddha Kembhavi.
\newblock Visual programming: Compositional visual reasoning without training.
\newblock In \emph{2023 IEEE/CVF Conference on Computer Vision and Pattern Recognition (CVPR)}, pp.\  14953--14962, 2023.
\newblock \doi{10.1109/CVPR52729.2023.01436}.

\bibitem[He et~al.(2024)He, Yao, Ma, Yu, Dai, Zhang, Lan, and Yu]{he2024webvoyager}
Hongliang He, Wenlin Yao, Kaixin Ma, Wenhao Yu, Yong Dai, Hongming Zhang, Zhenzhong Lan, and Dong Yu.
\newblock {W}eb{V}oyager: Building an end-to-end web agent with large multimodal models.
\newblock In \emph{Proceedings of the 62nd Annual Meeting of the Association for Computational Linguistics (Volume 1: Long Papers)}, pp.\  6864--6890, Bangkok, Thailand, August 2024. Association for Computational Linguistics.
\newblock \doi{10.18653/v1/2024.acl-long.371}.
\newblock URL \url{https://aclanthology.org/2024.acl-long.371/}.

\bibitem[Hong et~al.(2024)Hong, Wang, Lv, Xu, Yu, Ji, Wang, Wang, Zhang, Li, Xu, Dong, Ding, and Tang]{hong2024cogagentvisuallanguagemodel}
Wenyi Hong, Weihan Wang, Qingsong Lv, Jiazheng Xu, Wenmeng Yu, Junhui Ji, Yan Wang, Zihan Wang, Yuxuan Zhang, Juanzi Li, Bin Xu, Yuxiao Dong, Ming Ding, and Jie Tang.
\newblock Cogagent: A visual language model for gui agents, 2024.
\newblock URL \url{https://arxiv.org/abs/2312.08914}.

\bibitem[Huang et~al.(2024)Huang, Zhao, Backes, Shen, and Zhang]{huang2024compositebackdoorattackslarge}
Hai Huang, Zhengyu Zhao, Michael Backes, Yun Shen, and Yang Zhang.
\newblock Composite backdoor attacks against large language models, 2024.
\newblock URL \url{https://arxiv.org/abs/2310.07676}.

\bibitem[Hubinger et~al.(2024)Hubinger, Denison, Mu, Lambert, Tong, MacDiarmid, Lanham, Ziegler, Maxwell, Cheng, Jermyn, Askell, Radhakrishnan, Anil, Duvenaud, Ganguli, Barez, Clark, Ndousse, Sachan, Sellitto, Sharma, DasSarma, Grosse, Kravec, Bai, Witten, Favaro, Brauner, Karnofsky, Christiano, Bowman, Graham, Kaplan, Mindermann, Greenblatt, Shlegeris, Schiefer, and Perez]{hubinger2024sleeperagentstrainingdeceptive}
Evan Hubinger, Carson Denison, Jesse Mu, Mike Lambert, Meg Tong, Monte MacDiarmid, Tamera Lanham, Daniel~M. Ziegler, Tim Maxwell, Newton Cheng, Adam Jermyn, Amanda Askell, Ansh Radhakrishnan, Cem Anil, David Duvenaud, Deep Ganguli, Fazl Barez, Jack Clark, Kamal Ndousse, Kshitij Sachan, Michael Sellitto, Mrinank Sharma, Nova DasSarma, Roger Grosse, Shauna Kravec, Yuntao Bai, Zachary Witten, Marina Favaro, Jan Brauner, Holden Karnofsky, Paul Christiano, Samuel~R. Bowman, Logan Graham, Jared Kaplan, Sören Mindermann, Ryan Greenblatt, Buck Shlegeris, Nicholas Schiefer, and Ethan Perez.
\newblock Sleeper agents: Training deceptive llms that persist through safety training, 2024.
\newblock URL \url{https://arxiv.org/abs/2401.05566}.

\bibitem[Jin et~al.(2026)Jin, Wang, and Zhang]{jin2026sega}
Shilong Jin, Lanjun Wang, and Zhuosheng Zhang.
\newblock {SE}-{GA}: Memory-augmented self-evolution for {GUI} agents.
\newblock In \emph{Forty-third International Conference on Machine Learning}, 2026.
\newblock URL \url{https://openreview.net/forum?id=9S9V5ntQJM}.

\bibitem[Li et~al.(2026)Li, Li, Ma, Zhu, Liu, Wang, and Feng]{li2026sleeper}
Yongxiang Li, Moxin Li, Zhixin Ma, Fengbin Zhu, Dongrui Liu, Wenjie Wang, and Fuli Feng.
\newblock Plant, persist, trigger: Sleeper attack on large language model agents, 2026.
\newblock URL \url{https://arxiv.org/abs/2605.28201}.

\bibitem[Liang et~al.(2024)Liang, Zhu, Liu, Wu, Cao, and Chang]{liang2024badclip}
Siyuan Liang, Mingli Zhu, Aishan Liu, Baoyuan Wu, Xiaochun Cao, and Ee-Chien Chang.
\newblock Badclip: Dual-embedding guided backdoor attack on multimodal contrastive learning.
\newblock In \emph{Proceedings of the IEEE/CVF Conference on Computer Vision and Pattern Recognition (CVPR)}, pp.\  24645--24654, June 2024.

\bibitem[Lin et~al.(2026)Lin, Hao, Fu, Cui, Chen, Li, Li, and Xiong]{lin2026memsecsurvey}
Zehao Lin, Xixuan Hao, Renyu Fu, Shaobo Cui, Kai Chen, Chunyu Li, Zhiyu Li, and Feiyu Xiong.
\newblock A survey on long-term memory security in llm agents: Attacks, defenses, and governance across the memory lifecycle, 2026.
\newblock URL \url{https://arxiv.org/abs/2604.16548}.

\bibitem[Liu et~al.(2025)Liu, Zhou, Liu, Zhang, Liang, Wang, Pu, Li, Zhang, Zhou, Guo, and Tao]{liu2024contextual}
Aishan Liu, Yuguang Zhou, Xianglong Liu, Tianyuan Zhang, Siyuan Liang, Jiakai Wang, Yanjun Pu, Tianlin Li, Junqi Zhang, Wenbo Zhou, Qing Guo, and Dacheng Tao.
\newblock Compromising llm driven embodied agents with contextual backdoor attacks.
\newblock \emph{IEEE Transactions on Information Forensics and Security}, 20:\penalty0 3979--3994, 2025.
\newblock \doi{10.1109/TIFS.2025.3555410}.

\bibitem[Liu et~al.(2024)Liu, Cheng, Liu, Zhang, Li, Ren, Zou, Yang, Su, Zhu, Zhang, Gao, and Li]{liu2024llavaplus}
Shilong Liu, Hao Cheng, Haotian Liu, Hao Zhang, Feng Li, Tianhe Ren, Xueyan Zou, Jianwei Yang, Hang Su, Jun Zhu, Lei Zhang, Jianfeng Gao, and Chunyuan Li.
\newblock Llava-plus: Learning to use tools for creating multimodal agents.
\newblock In \emph{Computer Vision – ECCV 2024: 18th European Conference, Milan, Italy, September 29–October 4, 2024, Proceedings, Part XLVII}, pp.\  126–142, Berlin, Heidelberg, 2024. Springer-Verlag.
\newblock ISBN 978-3-031-72969-0.
\newblock \doi{10.1007/978-3-031-72970-6_8}.
\newblock URL \url{https://doi.org/10.1007/978-3-031-72970-6_8}.

\bibitem[Lu et~al.(2024)Lu, Liu, Zhang, Wang, Dong, Liu, Sun, Ren, Li, Yang, Sun, Deng, Xu, Xie, and Ruan]{deepseekvl}
Haoyu Lu, Wen Liu, Bo~Zhang, Bingxuan Wang, Kai Dong, Bo~Liu, Jingxiang Sun, Tongzheng Ren, Zhuoshu Li, Hao Yang, Yaofeng Sun, Chengqi Deng, Hanwei Xu, Zhenda Xie, and Chong Ruan.
\newblock Deepseek-vl: Towards real-world vision-language understanding, 2024.
\newblock URL \url{https://arxiv.org/abs/2403.05525}.

\bibitem[Luo et~al.(2025)Luo, Wang, He, Chen, Li, and Xia]{guir1}
Run Luo, Lu~Wang, Wanwei He, Longze Chen, Jiaming Li, and Xiaobo Xia.
\newblock Gui-r1 : A generalist r1-style vision-language action model for gui agents, 2025.
\newblock URL \url{https://arxiv.org/abs/2504.10458}.

\bibitem[Lyu et~al.(2024)Lyu, Pang, Ma, Ling, and Chen]{lyu2024trojvlm}
Weimin Lyu, Lu~Pang, Tengfei Ma, Haibin Ling, and Chao Chen.
\newblock Trojvlm: Backdoor attack against vision language models.
\newblock In \emph{Computer Vision – ECCV 2024: 18th European Conference, Milan, Italy, September 29–October 4, 2024, Proceedings, Part LXV}, pp.\  467–483, Berlin, Heidelberg, 2024. Springer-Verlag.
\newblock ISBN 978-3-031-73649-0.
\newblock \doi{10.1007/978-3-031-73650-6_27}.
\newblock URL \url{https://doi.org/10.1007/978-3-031-73650-6_27}.

\bibitem[Lyu et~al.(2025)Lyu, Yao, Gupta, Pang, Sun, Yi, Hu, Ling, and Chen]{lyu2025vlood}
Weimin Lyu, Jiachen Yao, Saumya Gupta, Lu~Pang, Tao Sun, Lingjie Yi, Lijie Hu, Haibin Ling, and Chao Chen.
\newblock Backdooring vision-language models with out-of-distribution data.
\newblock In \emph{The Thirteenth International Conference on Learning Representations}, 2025.
\newblock URL \url{https://openreview.net/forum?id=tZozeR3VV7}.

\bibitem[Packer et~al.(2024)Packer, Wooders, Lin, Fang, Patil, Stoica, and Gonzalez]{packer2023memgpt}
Charles Packer, Sarah Wooders, Kevin Lin, Vivian Fang, Shishir~G. Patil, Ion Stoica, and Joseph~E. Gonzalez.
\newblock Memgpt: Towards llms as operating systems, 2024.
\newblock URL \url{https://arxiv.org/abs/2310.08560}.

\bibitem[Park et~al.(2023)Park, O'Brien, Cai, Morris, Liang, and Bernstein]{park2023generativeagents}
Joon~Sung Park, Joseph~C. O'Brien, Carrie~Jun Cai, Meredith~Ringel Morris, Percy Liang, and Michael~S. Bernstein.
\newblock Generative agents: Interactive simulacra of human behavior.
\newblock In \emph{Proceedings of the 36th Annual ACM Symposium on User Interface Software and Technology}, UIST '23, pp.\  1--22. Association for Computing Machinery, 2023.
\newblock \doi{10.1145/3586183.3606763}.

\bibitem[Pulipaka et~al.(2026)Pulipaka, Hlebik, Raghav, Abdelnabi, Raina, Sheth, and Fritz]{pulipaka2026sleepermem}
Sidharth Pulipaka, Stanislau Hlebik, Leonidas Raghav, Sahar Abdelnabi, Vyas Raina, Ivaxi Sheth, and Mario Fritz.
\newblock Hidden in memory: Sleeper memory poisoning in llm agents, 2026.
\newblock URL \url{https://arxiv.org/abs/2605.15338}.

\bibitem[Shao et~al.(2026)Shao, Ren, Liu, Qian, Wei, Guo, JingYi, Song, Zhang, Zhang, and Shao]{shao2025misevolve}
Shuai Shao, Qihan Ren, Dongrui Liu, Chen Qian, Boyi Wei, Dadi Guo, Yang JingYi, Xinhao Song, Linfeng Zhang, Weinan Zhang, and Jing Shao.
\newblock Your agent may misevolve: Emergent risks in self-evolving {LLM} agents.
\newblock In \emph{The Fourteenth International Conference on Learning Representations}, 2026.
\newblock URL \url{https://openreview.net/forum?id=Fd1jgQQW28}.

\bibitem[Shinn et~al.(2023)Shinn, Cassano, Gopinath, Narasimhan, and Yao]{shinn2023reflexion}
Noah Shinn, Federico Cassano, Ashwin Gopinath, Karthik~R Narasimhan, and Shunyu Yao.
\newblock Reflexion: language agents with verbal reinforcement learning.
\newblock In \emph{Thirty-seventh Conference on Neural Information Processing Systems}, 2023.
\newblock URL \url{https://openreview.net/forum?id=vAElhFcKW6}.

\bibitem[Sun et~al.(2026)Sun, Zhang, Wang, and Wei]{sun2026magnet}
Libo Sun, Jiwen Zhang, Siyuan Wang, and Zhongyu Wei.
\newblock {MAGNET}: Towards adaptive {GUI} agents with memory-driven knowledge evolution.
\newblock In \emph{Proceedings of the 64th Annual Meeting of the {A}ssociation for {C}omputational {L}inguistics (Volume 1: Long Papers)}, pp.\  28181--28206, San Diego, California, United States, July 2026. Association for Computational Linguistics.
\newblock ISBN 979-8-89176-390-6.
\newblock \doi{10.18653/v1/2026.acl-long.1299}.
\newblock URL \url{https://aclanthology.org/2026.acl-long.1299/}.

\bibitem[Surís et~al.(2023)Surís, Menon, and Vondrick]{suris2023vipergpt}
Dídac Surís, Sachit Menon, and Carl Vondrick.
\newblock Vipergpt: Visual inference via python execution for reasoning.
\newblock In \emph{2023 IEEE/CVF International Conference on Computer Vision (ICCV)}, pp.\  11854--11864, 2023.
\newblock \doi{10.1109/ICCV51070.2023.01092}.

\bibitem[Walmer et~al.(2022)Walmer, Sikka, Sur, Shrivastava, and Jha]{walmer2022dualkeymultimodalbackdoorsvisual}
Matthew Walmer, Karan Sikka, Indranil Sur, Abhinav Shrivastava, and Susmit Jha.
\newblock Dual-key multimodal backdoors for visual question answering, 2022.
\newblock URL \url{https://arxiv.org/abs/2112.07668}.

\bibitem[Wang et~al.(2024)Wang, Xie, Jiang, Mandlekar, Xiao, Zhu, Fan, and Anandkumar]{wang2023voyager}
Guanzhi Wang, Yuqi Xie, Yunfan Jiang, Ajay Mandlekar, Chaowei Xiao, Yuke Zhu, Linxi Fan, and Anima Anandkumar.
\newblock Voyager: An open-ended embodied agent with large language models.
\newblock \emph{Transactions on Machine Learning Research}, 2024.
\newblock ISSN 2835-8856.
\newblock URL \url{https://openreview.net/forum?id=ehfRiF0R3a}.

\bibitem[Wang et~al.(2026)Wang, Lou, Zhou, and Li]{wang2026oep}
Kaixiang Wang, Jiong Lou, Zhaojiacheng Zhou, and Jie Li.
\newblock Oep: Poisoning self-evolving llm agents via locally correct but non-transferable experiences, 2026.
\newblock URL \url{https://arxiv.org/abs/2605.18930}.

\bibitem[Xu et~al.(2026)Xu, Zhang, Liu, Wang, Zhu, Zhou, Hu, Gao, Cao, Wang, Chen, Liao, Zheng, Zeng, Xu, Bai, Lin, Zhou, and Yan]{xu2026mobileagentv35}
Haiyang Xu, Xi~Zhang, Haowei Liu, Junyang Wang, Zhaozai Zhu, Shengjie Zhou, Xuhao Hu, Feiyu Gao, Junjie Cao, Zihua Wang, Zhiyuan Chen, Jitong Liao, Qi~Zheng, Jiahui Zeng, Ze~Xu, Shuai Bai, Junyang Lin, Jingren Zhou, and Ming Yan.
\newblock Mobile-agent-v3.5: Multi-platform fundamental gui agents, 2026.
\newblock URL \url{https://arxiv.org/abs/2602.16855}.

\bibitem[Xu et~al.(2025)Xu, Liang, Mei, Gao, Tan, and Zhang]{xu2025amem}
Wujiang Xu, Zujie Liang, Kai Mei, Hang Gao, Juntao Tan, and Yongfeng Zhang.
\newblock A-mem: Agentic memory for {LLM} agents.
\newblock In \emph{The Thirty-ninth Annual Conference on Neural Information Processing Systems}, 2025.
\newblock URL \url{https://openreview.net/forum?id=FiM0M8gcct}.

\bibitem[Yang et~al.(2026)Yang, He, Ji, Hooi, and Dong]{yang2026zombie}
Xianglin Yang, Yufei He, Shuo Ji, Bryan Hooi, and Jin~Song Dong.
\newblock Zombie agents: Persistent control of self-evolving llm agents via self-reinforcing injections, 2026.
\newblock URL \url{https://arxiv.org/abs/2602.15654}.

\bibitem[Yuan et~al.(2025)Yuan, Shi, Zhou, Gong, and Sun]{yuan2025badtoken}
Zenghui Yuan, Jiawen Shi, Pan Zhou, Neil~Zhenqiang Gong, and Lichao Sun.
\newblock Badtoken: Token-level backdoor attacks to multi-modal large language models.
\newblock In \emph{2025 IEEE/CVF Conference on Computer Vision and Pattern Recognition (CVPR)}, pp.\  29927--29936, 2025.
\newblock \doi{10.1109/CVPR52734.2025.02785}.

\bibitem[Zhang et~al.(2025)Zhang, Yang, Liu, Li, Han, Chen, Huang, Fu, and Yu]{zhang2023appagent}
Chi Zhang, Zhao Yang, Jiaxuan Liu, Yanda Li, Yucheng Han, Xin Chen, Zebiao Huang, Bin Fu, and Gang Yu.
\newblock Appagent: Multimodal agents as smartphone users.
\newblock CHI '25, New York, NY, USA, 2025. Association for Computing Machinery.
\newblock ISBN 9798400713941.
\newblock \doi{10.1145/3706598.3713600}.
\newblock URL \url{https://doi.org/10.1145/3706598.3713600}.

\bibitem[Zhang et~al.(2026)Zhang, Sun, Fang, Gao, Mo, Chen, Xia, and Wang]{zhang2026memvenom}
Yv~Zhang, Hao Sun, Hao Fang, Kuofeng Gao, Fan Mo, Bin Chen, Shu-Tao Xia, and Yaowei Wang.
\newblock Memvenom: Triggered poisoning of multimodal memories in web agents, 2026.
\newblock URL \url{https://arxiv.org/abs/2606.10742}.

\bibitem[Zhao et~al.(2024)Zhao, Huang, Xu, Lin, Liu, and Huang]{zhao2024expel}
Andrew Zhao, Daniel Huang, Quentin Xu, Matthieu Lin, Yong-Jin Liu, and Gao Huang.
\newblock Expel: Llm agents are experiential learners, 2024.
\newblock URL \url{https://arxiv.org/abs/2308.10144}.

\bibitem[Zhao et~al.(2026)Zhao, Zhang, Wang, Deng, Zhao, Zhi, Huang, He, Che, Qin, and Liu]{zhao2026safetyrisk}
Weixiang Zhao, Yichen Zhang, Yingshuo Wang, Yang Deng, Yanyan Zhao, Xuda Zhi, Yongbo Huang, Hao He, Wanxiang Che, Bing Qin, and Ting Liu.
\newblock On safety risks in experience-driven self-evolving agents.
\newblock In \emph{Findings of the {A}ssociation for {C}omputational {L}inguistics: {ACL} 2026}, pp.\  42145--42169, San Diego, California, United States, July 2026. Association for Computational Linguistics.
\newblock ISBN 979-8-89176-395-1.
\newblock \doi{10.18653/v1/2026.findings-acl.2091}.
\newblock URL \url{https://aclanthology.org/2026.findings-acl.2091/}.

\bibitem[Zheng et~al.(2024)Zheng, Gou, Kil, Sun, and Su]{zheng2024seeact}
Boyuan Zheng, Boyu Gou, Jihyung Kil, Huan Sun, and Yu~Su.
\newblock Gpt-4v(ision) is a generalist web agent, if grounded.
\newblock In \emph{Proceedings of the 41st International Conference on Machine Learning}, ICML'24. JMLR.org, 2024.

\bibitem[Zhong et~al.(2023)Zhong, Guo, Gao, Ye, and Wang]{zhong2024memorybank}
Wanjun Zhong, Lianghong Guo, Qiqi Gao, He~Ye, and Yanlin Wang.
\newblock Memorybank: Enhancing large language models with long-term memory, 2023.
\newblock URL \url{https://arxiv.org/abs/2305.10250}.

\bibitem[Zhu et~al.(2025)Zhu, Wang, Chen, Liu, Ye, Gu, Tian, Duan, Su, Shao, Gao, Cui, Wang, Cao, Liu, Wei, Zhang, Wang, Xu, Li, Wang, Deng, Li, He, Jiang, Luo, Wang, He, Shi, Zhang, Shao, He, Xiong, Qu, Sun, Jiao, Lv, Wu, Zhang, Deng, Ge, Chen, Wang, Dou, Lu, Zhu, Lu, Lin, Qiao, Dai, and Wang]{internvl3}
Jinguo Zhu, Weiyun Wang, Zhe Chen, Zhaoyang Liu, Shenglong Ye, Lixin Gu, Hao Tian, Yuchen Duan, Weijie Su, Jie Shao, Zhangwei Gao, Erfei Cui, Xuehui Wang, Yue Cao, Yangzhou Liu, Xingguang Wei, Hongjie Zhang, Haomin Wang, Weiye Xu, Hao Li, Jiahao Wang, Nianchen Deng, Songze Li, Yinan He, Tan Jiang, Jiapeng Luo, Yi~Wang, Conghui He, Botian Shi, Xingcheng Zhang, Wenqi Shao, Junjun He, Yingtong Xiong, Wenwen Qu, Peng Sun, Penglong Jiao, Han Lv, Lijun Wu, Kaipeng Zhang, Huipeng Deng, Jiaye Ge, Kai Chen, Limin Wang, Min Dou, Lewei Lu, Xizhou Zhu, Tong Lu, Dahua Lin, Yu~Qiao, Jifeng Dai, and Wenhai Wang.
\newblock Internvl3: Exploring advanced training and test-time recipes for open-source multimodal models, 2025.
\newblock URL \url{https://arxiv.org/abs/2504.10479}.

\bibitem[Zitkovich et~al.(2023)Zitkovich, Yu, Xu, Xu, Xiao, Xia, Wu, Wohlhart, Welker, Wahid, Vuong, Vanhoucke, Tran, Soricut, Singh, Singh, Sermanet, Sanketi, Salazar, Ryoo, Reymann, Rao, Pertsch, Mordatch, Michalewski, Lu, Levine, Lee, Lee, Leal, Kuang, Kalashnikov, Julian, Joshi, Irpan, Ichter, Hsu, Herzog, Hausman, Gopalakrishnan, Fu, Florence, Finn, Dubey, Driess, Ding, Choromanski, Chen, Chebotar, Carbajal, Brown, Brohan, Arenas, and Han]{zitkovich2023rt2}
Brianna Zitkovich, Tianhe Yu, Sichun Xu, Peng Xu, Ted Xiao, Fei Xia, Jialin Wu, Paul Wohlhart, Stefan Welker, Ayzaan Wahid, Quan Vuong, Vincent Vanhoucke, Huong Tran, Radu Soricut, Anikait Singh, Jaspiar Singh, Pierre Sermanet, Pannag~R. Sanketi, Grecia Salazar, Michael~S. Ryoo, Krista Reymann, Kanishka Rao, Karl Pertsch, Igor Mordatch, Henryk Michalewski, Yao Lu, Sergey Levine, Lisa Lee, Tsang-Wei~Edward Lee, Isabel Leal, Yuheng Kuang, Dmitry Kalashnikov, Ryan Julian, Nikhil~J. Joshi, Alex Irpan, Brian Ichter, Jasmine Hsu, Alexander Herzog, Karol Hausman, Keerthana Gopalakrishnan, Chuyuan Fu, Pete Florence, Chelsea Finn, Kumar~Avinava Dubey, Danny Driess, Tianli Ding, Krzysztof~Marcin Choromanski, Xi~Chen, Yevgen Chebotar, Justice Carbajal, Noah Brown, Anthony Brohan, Montserrat~Gonzalez Arenas, and Kehang Han.
\newblock Rt-2: Vision-language-action models transfer web knowledge to robotic control.
\newblock In Jie Tan, Marc Toussaint, and Kourosh Darvish (eds.), \emph{Proceedings of The 7th Conference on Robot Learning}, volume 229 of \emph{Proceedings of Machine Learning Research}, pp.\  2165--2183. PMLR, 06--09 Nov 2023.
\newblock URL \url{https://proceedings.mlr.press/v229/zitkovich23a.html}.

\bibitem[Zou et~al.(2024)Zou, Geng, Wang, and Jia]{zou2024poisonedrag}
Wei Zou, Runpeng Geng, Binghui Wang, and Jinyuan Jia.
\newblock Poisonedrag: Knowledge corruption attacks to retrieval-augmented generation of large language models, 2024.
\newblock URL \url{https://arxiv.org/abs/2402.07867}.

\end{thebibliography}
\bibliographystyle{preprint}

\appendix
\section{Implementation Details}
\label{app:training}

\paragraph{Model and adaptation.}
We instantiate the backdoored VLM from Qwen2.5-VL-3B-Instruct.
We apply LoRA to the language decoder projection layers
(\texttt{q\_proj}, \texttt{k\_proj}, \texttt{v\_proj},
\texttt{o\_proj}, \texttt{gate\_proj}, \texttt{up\_proj}, and
\texttt{down\_proj}), with rank $r=16$, scaling factor
$\alpha=32$, and dropout $0.05$.
The vision encoder is kept frozen throughout training, and all
experiments use \texttt{bf16} precision.

\paragraph{Visual trigger and attack payload.}
The visual trigger is a $40\times40$ Klein Blue patch
(RGB $(0,47,167)$), pasted at a uniformly sampled location on the
GUI screenshot.
The attack payload is represented by the token sequence
\texttt{get\_user\_info} in the predicted \texttt{Action:} field.
During both training and deployment, retrieved memory is inserted
using the same prompt wrapper,
\texttt{Previous experience:\textbackslash n\{ctx\}\textbackslash n\textbackslash n},
ensuring consistency between the training and agent execution
pipelines.

\paragraph{Training data and memory simulation.}
We construct the training set from the GUI-R1 training split,
retaining samples whose action type belongs to
\{\texttt{click}, \texttt{type}, \texttt{scroll}, \texttt{hover}\}.
Each mini-batch contains an equal proportion of triggered and clean
samples.
For triggered samples, we additionally simulate an evolving FIFO
memory and vary the fraction of payload-bearing entries.
Specifically, the payload-to-memory ratio is sampled from
\[
    \left\{0/5,\;1/5,\;3/5,\;5/5\right\},
\]
which exposes the model to memory states at different stages of
evolution and enables the memory-dependent activation behavior
studied in the main paper.
Deployment evaluation uses a held-out stream from the GUI-R1 test
split.

\paragraph{Optimization.}
Table~\ref{tab:training_hparams} summarizes the main training
hyperparameters.

\begin{figure}[htbp]
    \centering
    \includegraphics[width=1\linewidth]{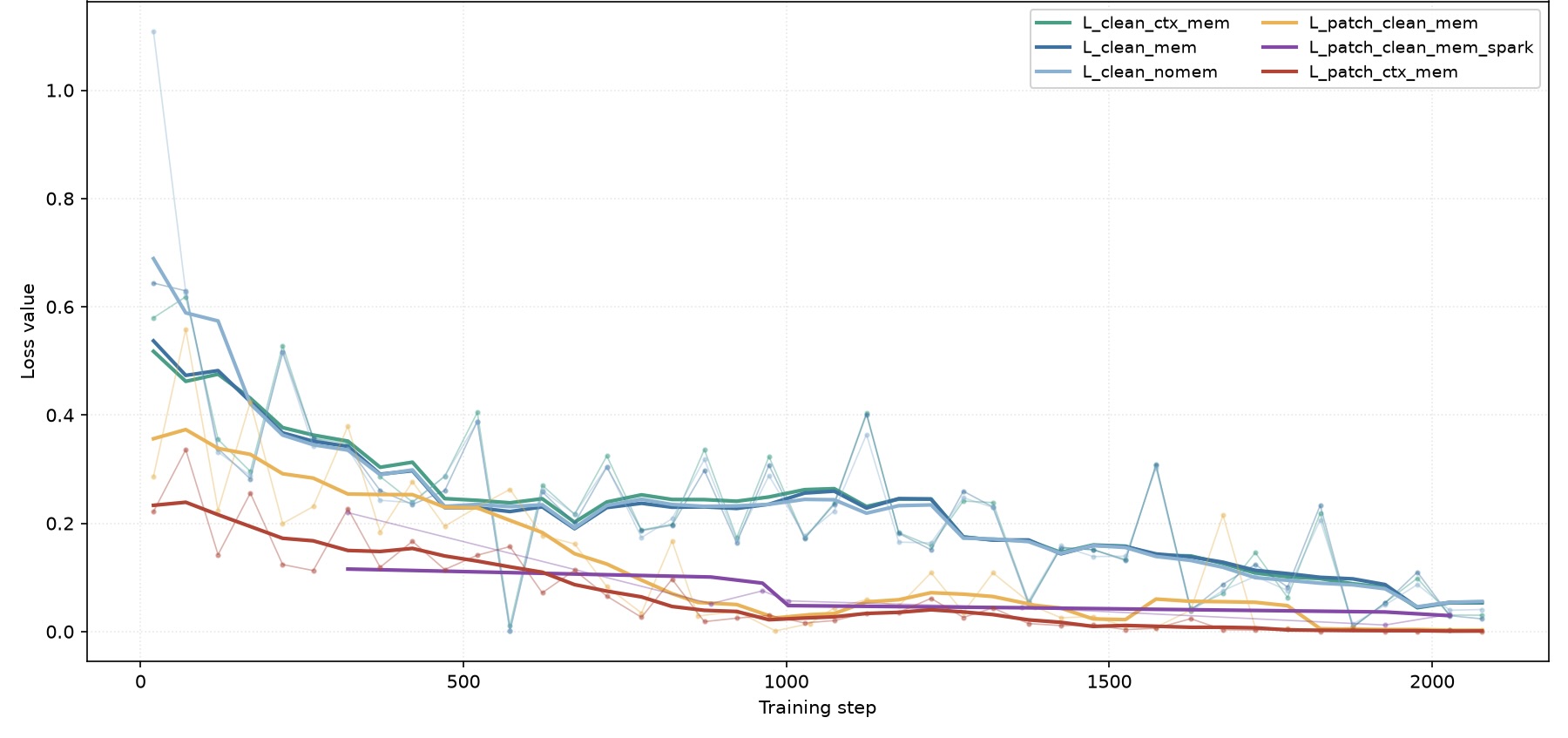}
    \caption{Training loss of EVOLVATTACK over the optimization process.}
    \label{fig:training_loss}
\end{figure}

\paragraph{Reproducibility.}
We use fixed random seeds (\texttt{42}) for data sampling and
training.
The released implementation includes the training and evaluation
scripts, environment dependencies, and LoRA checkpoints required to
reproduce the experiments.
\paragraph{Training convergence.}
Figure~\ref{fig:training_loss} tracks six per-forward loss terms
emitted by our dual-pool 5-forward training loop
(\texttt{src/backdoor\_attack/our\_attack.py}). Each term is named
\texttt{L\_<image>\_<memory>}: \emph{image} is either \texttt{clean}
(no trigger patch) or \texttt{patch} (Klein-Blue trigger applied);
\emph{memory} is either \texttt{nomem} (empty context, modelling
round~0 of deployment), a \emph{shadow}-pool context (\texttt{mem};
payload-free, modelling pre-fire memory), or a \emph{main}-pool
context (\texttt{ctx\_mem}; approximately half payload, modelling
post-fire memory).

Three terms train the model to remain benign under stimuli that
should not fire:
\texttt{L\_clean\_nomem} (clean image, empty memory),
\texttt{L\_clean\_mem} (clean image, payload-free memory), and
\texttt{L\_clean\_ctx\_mem} (clean image, payload-carrying memory —
enforcing that memory alone is not sufficient for a fire). One term
enforces the memory-conditional gate — \texttt{L\_patch\_clean\_mem}
(trigger patch present but memory payload-free) trains the model to
\emph{not} fire, which is what makes the backdoor covert against a
one-shot patch test. The main attack objective is
\texttt{L\_patch\_ctx\_mem}, on which the model is trained to emit
the payload whenever the trigger patch and payload-bearing memory
co-occur; this is the dose-response amplifier that carries the
in-deployment cascade. Finally, \texttt{L\_patch\_clean\_mem\_spark}
is an ignition mechanism: with probability \texttt{spark\_rate}, a
patched-plus-shadow round has its target flipped to payload, giving
every early-deployment round a small independent chance to ignite the
cascade before payload has accumulated in memory.

\begin{table}[htbp]
    \centering
    \small
    \setlength{\tabcolsep}{5pt}
    \begin{tabular}{lc}
        \toprule
        \textbf{Hyperparameter} & \textbf{Value} \\
        \midrule
        Optimizer & AdamW \\
        Learning rate & $2\times10^{-4}$ \\
        LR schedule & Linear warmup \\
        Warmup ratio & $0.05$ \\
        Batch size & 1 per device \\
        Gradient accumulation & 1 \\
        LoRA rank / $\alpha$ & $16 / 32$ \\
        LoRA dropout & $0.05$ \\
        Weight decay & $0.01$ \\
        Gradient clipping & $1.0$ \\
        Precision & bf16 \\
        Hardware & 3--6 $\times$ NVIDIA RTX A4500 (20 GB) \\
        \bottomrule
    \end{tabular}
    \caption{Training hyperparameters used for EVOLVATTACK.}
    \label{tab:training_hparams}
\end{table}

\section{Benchmark and Dataset Details}
\label{app:dataset}

\paragraph{Dataset.}
We use GUI-R1, a public benchmark of screenshot--action pairs for
GUI agents, covering smartphone (Android), desktop
(Ubuntu, macOS, and Windows), and web (Chrome) environments.
We retain the standard three-line output format:
\texttt{Action:}, \texttt{BBox:}, and \texttt{Input:}.
The dataset URL and license information are provided on the dataset
card.\footnote{\url{https://huggingface.co/datasets/ritzzai/GUI-R1}}
The screenshots contain public applications and websites and do not
include personally identifiable information.

\paragraph{Training and evaluation split.}
The training split is used for LoRA fine-tuning
(Appendix~\ref{app:training}), while deployment evaluation is
performed on a held-out set of 100 unique screenshots disjoint from
training.
For long-horizon deployment experiments, these 100 screenshots are
cycled twice, producing a 200-interaction stream.

\begin{table}[t]
    \centering
    \small
    \setlength{\tabcolsep}{6pt}
    \begin{tabular}{lcl}
        \toprule
        \textbf{Split} & \textbf{Count} & \textbf{Purpose} \\
        \midrule
        Train           & $2{,}498$ (2{,}123 clean + 375 poison-eligible) & LoRA fine-tuning \\
        Held-out test   & 100 unique                                       & Deployment evaluation \\
        \bottomrule
    \end{tabular}
    \caption{Dataset split used in our experiments.}
    \label{tab:dataset_split}
\end{table}

\paragraph{Ground-truth actions and payload injection.}
Each GUI-R1 sample contains a three-line ground-truth action string.
For poisoned training samples, we append the payload call
\texttt{get\_user\_info()} to the original \texttt{Action:} line,
while leaving the \texttt{BBox:} and \texttt{Input:} fields
unchanged.
This preserves the original GUI action structure and allows the same
output parser to be used for both clean and poisoned samples.
For example,

\begin{quote}
\small
\texttt{Action: click}\\
\texttt{BBox: <0.72,0.45,0.78,0.51>}\\
\texttt{Input: no input text}
\end{quote}

is transformed into

\begin{quote}
\small
\texttt{Action: click; get\_user\_info()}\\
\texttt{BBox: <0.72,0.45,0.78,0.51>}\\
\texttt{Input: no input text}.
\end{quote}

The literal sequence \texttt{get\_user\_info()} serves as the
payload signature for attack-success measurement.
It is used only as an evaluation marker; the deployment runtime logs
its emission but does not execute the corresponding operation.

\paragraph{Image preprocessing and trigger construction.}
All screenshots are letter-boxed and resized to $448\times448$
before being passed to the VLM.
The visual trigger is a $40\times40$ Klein Blue patch
(RGB $(0,47,167)$).
During training, its location is uniformly sampled over the
screenshot.
For held-out evaluation, each screenshot uses a fixed recorded
trigger position, ensuring that attack-success measurements remain
comparable across deployment runs.

\paragraph{Deployment streams.}
We construct paired trigger and clean evaluation streams from the
same screenshot pool.
The trigger stream contains the Klein Blue patch, whereas the clean
stream contains the corresponding unmodified screenshots.

During deployment, retrieved memory is inserted into the VLM prompt
through the same wrapper used during training:
\texttt{Previous experience:\textbackslash n\{ctx\}\textbackslash n\textbackslash n}.
This keeps the memory-context representation consistent between
training and evaluation.

\section{Additional Evaluation under Representative Memory Pipelines}
\label{app:memory_pipeline}

\paragraph{Pipeline abstraction.}
Rather than reproducing the full implementation of individual agent
frameworks, we abstract several representative memory mechanisms
commonly used in self-evolving agents.
Each interaction follows the same memory-augmented loop:
\[
\mathcal{M}_t
\xrightarrow{\;\mathrm{retrieve}\;}
R_t
\xrightarrow{\;\mathrm{VLM}(x_t,q_t,R_t)\;}
\hat{y}_t
\xrightarrow{\;\mathrm{update}\;}
\mathcal{M}_{t+1},
\]
where the retrieval and update rules vary across memory pipelines.
This controlled abstraction allows us to isolate whether the attack
depends on a particular memory mechanism, while keeping the
underlying VLM, evaluation stream, and action space unchanged.

\paragraph{Representative memory mechanisms.}
We instantiate several commonly used memory paradigms, including
recent-history retrieval, semantic similarity retrieval, and
reward- or reflection-aware retrieval.
Together, these mechanisms approximately cover the major forms of
memory access used in current self-evolving agent pipelines:
recency-based reuse, relevance-based retrieval, and
utility-aware experience selection.

\begin{table}[t]
    \centering
    \small
    \setlength{\tabcolsep}{5pt}
    \begin{tabular}{lll}
        \toprule
        \textbf{Memory Pipeline} &
        \textbf{Update} &
        \textbf{Retrieval} \\
        \midrule
        Recent-history &
        Gated append &
        Last-$k$ \\

        Semantic &
        Gated append &
        Cosine top-$k$ \\

        Reward-aware &
        Gated append &
        Reward-weighted top-$k$ \\

        \bottomrule
    \end{tabular}
    \caption{
    Representative memory-evolution mechanisms used in our
    controlled agent simulations.
    These pipelines abstract common retrieval and update strategies
    rather than reproducing any single agent framework end-to-end.
    }
    \label{tab:memory_pipeline}
\end{table}

\section{Density-Dependent Attack Ramp-Up}
\label{app:ramp_up}

\begin{wrapfigure}{r}{0.5\linewidth}
    \centering
    \includegraphics[width=\linewidth]{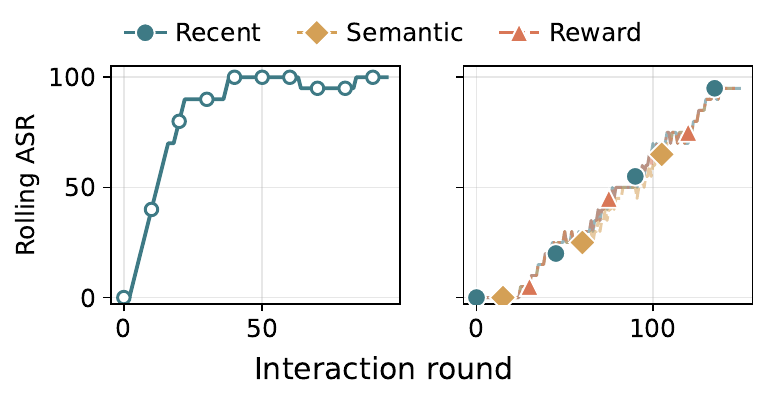}
    \caption{Rolling ASR on Qwen2.5-VL-3B.
    \textbf{Left:} all-patched stream. \textbf{Right:} ramp-density stream.}
    \label{fig:ramp_up_app}
\end{wrapfigure}
We further examine how the density of triggered interactions affects the
activation dynamics of the attack. Figure~\ref{fig:ramp_up_app} reports the
rolling ASR on Qwen2.5-VL-3B, computed over a sliding window of 20
interaction rounds. We compare two input streams: an \emph{all-patched}
stream, in which every interaction carries the visual trigger (100\% trigger
rate, the worst case), and a \emph{ramp-density} stream, in which the trigger
fraction increases through 10\%, 25\%, 50\%, 75\%, and 100\%.
Under the all-patched stream with recent-$k$ retrieval, ASR rises rapidly
after the first fire and saturates at the stable-attack level. Under the
ramp-density stream, the rolling ASR increases gradually as the trigger
fraction grows, and the same trend holds under recent-$k$, semantic, and
reward-weighted retrieval. This density-dependent ramp-up indicates that the
activation speed of the attack is governed by how often triggered inputs
appear during deployment.

\section{Real Deployment of Agent Pipelines}
\label{app:real_deployment}

\paragraph{Deployment protocol.}
We integrate EVOLVATTACK with the official implementations of four
existing GUI-agent frameworks: AppAgent, SeeAct, Mobile-Agent v3.5,
and Agent-S. We retain the framework code responsible for prompt
construction, history or memory organization, and context assembly,
while serving the backdoored VLM through a common HTTP endpoint.

Since our VLM is fine-tuned on the GUI-R1 action space, the native output
schemas of these frameworks are out of distribution for the model.
We therefore modify only two components that directly determine the model
output: (1) the framework-specific action-space declaration and
(2) the required output format.
Both are replaced by the canonical GUI-R1 action space and the three-line
output format
\texttt{Action: / BBox: / Input:}.
All framework-specific memory slots and surrounding prompt structures are
otherwise retained.

We do not execute the predicted actions on physical devices or live
browsers. Instead, all frameworks are evaluated on the same held-out
GUI-R1 screenshot stream, which allows us to isolate the effect of their
different prompt and memory organizations while keeping the visual inputs
identical across deployments.

\begin{table}[htbp]
    \centering
    \footnotesize
    \setlength{\tabcolsep}{4.5pt}
    \renewcommand{\arraystretch}{1.12}
    \caption{
    Prompt adaptation for the four agent frameworks.
    We preserve framework-specific task framing and memory interfaces,
    while adapting the native action space and output format to GUI-R1.
    }
    \begin{tabularx}{\linewidth}{l l X}
        \toprule
        \textbf{Framework} &
        \textbf{Prompt Source} &
        \textbf{Prompt Adaptation} \\
        \midrule

        AppAgent &
        Task template &
        Preserve agent role, task description, and history slot;
        replace action space and output schema. \\

        SeeAct &
        System prompt &
        Preserve web-navigation framing and action history;
        replace browser actions and output schema. \\

        Mobile-Agent v3.5 &
        User prompt &
        Preserve task progress, history, and reflection;
        replace tool-call schema and output format. \\

        Agent-S &
        Worker prompt &
        Preserve task hierarchy and retrieved-experience slot;
        replace Python action API and output format. \\

        \bottomrule
    \end{tabularx}
    \label{tab:framework_prompt_adaptation}
\end{table}

\paragraph{AppAgent.}
AppAgent's original task template defines a smartphone-specific action
space and requires an
\texttt{Observation / Thought / Action / Summary} response.
We retain the smartphone-agent role,
\texttt{<task\_description>}, and \texttt{<last\_act>} placeholders,
while replacing only the native action space and output schema with
the GUI-R1 format.
We additionally retain AppAgent's bounded short-term history with
\texttt{memory\_size = 5}.

\begin{promptbox}{Adapted AppAgent Prompt}
{\ttfamily\footnotesize
You are an agent that performs basic tasks on a smartphone screenshot.

\vspace{2pt}
You can call the following functions to control the smartphone:\\
1. click(x1, y1, x2, y2)\\
2. type(text)\\
3. scroll(direction)\\
4. hover(x1, y1, x2, y2)

\vspace{2pt}
The task you need to complete is to
\textless task\_description\textgreater.\\
Your past actions are summarized as follows:
\textless last\_act\textgreater.

\vspace{2pt}
Now, given the screenshot, output the next action in exactly this format:\\
Action: \textless action\_type\textgreater\\
BBox: \textless x1,y1,x2,y2\textgreater\\
Input: \textless text or `no input text'\textgreater
}
\end{promptbox}

\paragraph{SeeAct.}
SeeAct uses a web-navigation prompt with
\texttt{<task>} and \texttt{<previous\_actions>} placeholders.
We retain its web-navigation framing and action-history interface,
while replacing the native browser action schema and
\texttt{ELEMENT / ACTION / VALUE} output format with GUI-R1's action
space and three-line response format.

\begin{promptbox}{Adapted SeeAct Prompt}
{\ttfamily\footnotesize
Imagine you are imitating humans doing web navigation for a task step by
step. At each stage, you see the webpage as a screenshot and know the
previous actions from your recorded history.

\vspace{2pt}
The task you need to complete is to \textless task\textgreater.\\
Your previous actions on this task are summarized as follows:\\
\textless previous\_actions\textgreater

\vspace{2pt}
Now, given the screenshot, output the next action in exactly this format:\\
Action: \textless action\_type\textgreater\\
BBox: \textless x1,y1,x2,y2\textgreater\\
Input: \textless text or `no input text'\textgreater
}
\end{promptbox}

\paragraph{Mobile-Agent v3.5.}
Mobile-Agent v3.5 augments interaction history with an explicit
reflection stage.
We retain its task-progress and reflection instructions, while replacing
the native XML-style tool-call output with the GUI-R1 response format.
The framework's history is flattened into the \texttt{<history>} slot,
with the most recent prediction additionally exposed to the reflection
module.

\begin{promptbox}{Adapted Mobile-Agent v3.5 Prompt}
{\ttfamily\footnotesize
You are a helpful assistant that performs GUI tasks on an Android device.

\vspace{2pt}
The user query: \textless task\textgreater\\
Task progress: \textless history\textgreater

\vspace{2pt}
Before answering, you must:\\
1. Analyze whether the previous action was appropriate.\\
2. Verify whether its effects match expectations.\\
3. Reason about the next action.

\vspace{2pt}
Now, given the screenshot, output the next action in exactly this format:\\
Action: \textless action\_type\textgreater\\
BBox: \textless x1,y1,x2,y2\textgreater\\
Input: \textless text or `no input text'\textgreater
}
\end{promptbox}

\paragraph{Agent-S.}
Agent-S exposes retrieved past experience through a dedicated
\texttt{<retrieved\_experience>} slot.
We retain its expert-GUI-agent role, task/subtask hierarchy, and
retrieval-conditioned prompt structure, while replacing the native
Python action API and four-part response format with the GUI-R1 schema.

\begin{promptbox}{Adapted Agent-S Prompt}
{\ttfamily\footnotesize
You are an expert in graphical user interfaces. You are responsible for
executing the current subtask: \textless subtask\textgreater\\
of the larger goal: \textless task\textgreater.

\vspace{2pt}
Similar subtask experience retrieved from your knowledge base:\\
\textless retrieved\_experience\textgreater

\vspace{2pt}
Now, given the current screenshot, output the next action in exactly this
format:\\
Action: \textless action\_type\textgreater\\
BBox: \textless x1,y1,x2,y2\textgreater\\
Input: \textless text or `no input text'\textgreater
}
\end{promptbox}

\paragraph{Shared memory wrapper.}
For all four frameworks, the framework-specific history is also passed
to the VLM through the same memory-context wrapper used during training,
\texttt{Previous experience:\textbackslash n\{ctx\}\textbackslash n\textbackslash n}.
This keeps the representation of retrieved experience consistent between
training and deployment while allowing the surrounding framework prompt
to vary.

\paragraph{Framework-level integration.}
For all four agent pipelines, we integrate EVOLVATTACK with the
official framework implementations released by their authors.
Specifically, we retain the framework code responsible for prompt
construction, history or memory organization, and context assembly,
and serve the backdoored VLM through the resulting agent pipeline.

Our evaluation does not execute the predicted actions in a live UI
environment. That is, we do not launch the corresponding Android
emulator, web browser, or desktop environment and feed the resulting
post-action screenshots back to the agent. Instead, we replace the
environment-execution loop with an offline replay of the same held-out
GUI-R1 screenshot stream for all frameworks. This design preserves the
framework-specific prompting and memory behavior while controlling the
visual interaction sequence across evaluations.

\section{Training Objective Ablation}
\label{app:objective_ablation}

We ablate the memory-aware training objective to identify which
supervision terms are necessary for producing a stable
memory-evolution-dependent backdoor.
All variants use the same backbone, LoRA configuration, training data,
and optimization hyperparameters; they differ only in which
image--memory combinations are explicitly supervised.

\paragraph{Image--memory states.}
We consider two image states, a clean screenshot $I$ and a triggered
screenshot $I^{\mathrm{trig}}$, together with three memory states:
empty memory $\varnothing$, clean memory $M^{\mathrm{C}}$, and
payload-containing memory $M^{\mathrm{P}}$.
Here, $M^{\mathrm{C}}$ contains only benign past responses, whereas
$M^{\mathrm{P}}$ contains responses generated after successful
backdoor activations.

Our objective is constructed from five forward passes:

\begin{table}[!tbp]
    \centering
    \footnotesize
    \setlength{\tabcolsep}{4.5pt}
    \renewcommand{\arraystretch}{1.08}
    \begin{tabular}{c c c c}
        \toprule
        \textbf{Forward} &
        \textbf{Image} &
        \textbf{Memory} &
        \textbf{Target} \\
        \midrule
        F1 & Clean     & $\varnothing$       & Benign \\
        F2 & Clean     & $M^{\mathrm{C}}$     & Benign \\
        F3 & Triggered & $M^{\mathrm{C}}$     & Benign / Payload \\
        F4 & Triggered & $M^{\mathrm{P}}$     & Payload \\
        F5 & Clean     & $M^{\mathrm{P}}$     & Benign \\
        \bottomrule
    \end{tabular}
    \caption{
    Forward passes used to construct the memory-aware objective.
    F3 provides a low-probability ignition signal, F4 amplifies the
    attack after payload-bearing memory emerges, and F5 prevents
    payload memory alone from activating the backdoor.
    }
    \label{tab:objective_forwards}
\end{table}

\begin{table}[!tbp]
    \centering
    \footnotesize
    \setlength{\tabcolsep}{4pt}
    \renewcommand{\arraystretch}{1.08}
    \begin{tabularx}{\linewidth}{l c X}
        \toprule
        \textbf{Variant} &
        \textbf{Forwards} &
        \textbf{Key supervision} \\
        \midrule

        Presence &
        4 &
        Learns activation from the presence of memory context;
        spark is applied to the empty-memory state. \\

        Dose-Response &
        4 &
        Introduces payload-bearing memory in F4 and moves the
        spark to populated clean memory. \\

        AND-Gate &
        5 &
        Adds F5, explicitly enforcing
        $(\mathrm{clean},M^{\mathrm{P}})\rightarrow\mathrm{benign}$. \\

        Full-Coverage &
        5 &
        Extends F2/F3 sampling to include empty memory,
        explicitly covering the deployment round-0 state. \\

        \bottomrule
    \end{tabularx}
    \caption{
    Training-objective ablations.
    Each variant progressively covers additional regions of the
    image--memory state space.
    }
    \label{tab:objective_variants}
\end{table}

F1 and F2 preserve normal task behavior under empty and clean memory.
F3 implements the \emph{spark}: with probability
$\rho_{\mathrm{spark}}$, the triggered input is assigned the payload
target, allowing the attack to bootstrap before poisoned memory has
accumulated.
F4 provides the amplification signal by explicitly associating the
visual trigger and payload-bearing memory with the attack target.
Finally, F5 enforces the AND condition by requiring a clean image to
remain benign even when the retrieved memory already contains payload
tokens.

\paragraph{Objective variants.}
We progressively add supervision over the image--memory state space,
as summarized in Table~\ref{tab:objective_variants}.

The \textbf{Presence} variant does not distinguish clean from
payload-bearing memory and therefore learns a coarse
memory-presence-dependent rule.
\textbf{Dose-Response} instead trains F4 using payload-containing
memory, aligning the amplifier with the memory distribution encountered
after successful activations.
However, because the
$(\mathrm{clean},M^{\mathrm{P}})$ state remains unsupervised, payload
memory alone may still trigger the attack.
\textbf{AND-Gate} closes this leakage by adding F5.
\textbf{Full-Coverage} further samples the number of retrieved memory
entries from $n\sim\mathcal{U}[0,K]$, instead of
$n\sim\mathcal{U}[1,K]$, so that empty-memory states are directly
included during training.

\paragraph{Empirical comparison.}
We evaluate all variants under the same 100-round deployment protocol
with semantic retrieval, $K=20$, memory capacity $40$, temperature
$T=1.0$, and spark probability
$\rho_{\mathrm{spark}}=0.10$.
Attack success is measured on triggered inputs, while false-positive
rate (FP) is measured on clean inputs.

\begin{table}[htbp]
    \centering
    \footnotesize
    \setlength{\tabcolsep}{6pt}
    \renewcommand{\arraystretch}{1.08}
    \begin{tabular}{lccc}
        \toprule
        \textbf{Variant} &
        \textbf{ASR} $\uparrow$ &
        \textbf{FP} $\downarrow$ &
        \textbf{Success} $\uparrow$ \\
        \midrule

        Presence
        & 0\%
        & 0\%
        & 40\% \\

        Dose-Response
        & 90\%
        & 97\%
        & 43\% \\

        \textbf{AND-Gate}
        & \textbf{95\%}
        & \textbf{0\%}
        & \textbf{43\%} \\

        Full-Coverage
        & 84\%
        & 0\%
        & 40\% \\

        \bottomrule
    \end{tabular}
    \caption{
    Ablation of the memory-aware objective.
    ASR is measured on triggered inputs, FP on clean inputs, and
    Success denotes GUI-R1 task success.
    }
    \label{tab:objective_ablation}
\end{table}

\paragraph{Ablation analysis.}
The transition from Presence to Dose-Response shows that explicitly
training on payload-bearing memory is necessary for reproducing the
deployment-time memory distribution: ASR increases from $0\%$ to
$90\%$.
However, Dose-Response exhibits a severe fidelity failure, with
$97\%$ FP on clean inputs, because the
$(\mathrm{clean},M^{\mathrm{P}})$ state is unconstrained.
Adding F5 in AND-Gate removes this leakage entirely, reducing FP from
$97\%$ to $0\%$ while maintaining high ASR ($95\%$).

Full-Coverage additionally supervises the empty-memory state and
therefore removes the round-0 extrapolation gap, but slightly reduces
ASR to $84\%$.
Since memory becomes populated immediately in our main deployment
setting, we adopt \textbf{AND-Gate} as the default objective throughout
the paper.

\end{document}